\documentclass[sigconf]{aamas}  

\AtBeginDocument{%
  \providecommand\BibTeX{{%
    \normalfont B\kern-0.5em{\scshape i\kern-0.25em b}\kern-0.8em\TeX}}}

\usepackage{booktabs}
\usepackage{flushend} 

\setcopyright{ifaamas}  
\acmDOI{}  
\acmISBN{}  
\acmConference[AAMAS'20]{Proc.\@ of the 19th International Conference on Autonomous Agents and Multiagent Systems (AAMAS 2020)}{May 9--13, 2020}{Auckland, New Zealand}{B.~An, N.~Yorke-Smith, A.~El~Fallah~Seghrouchni, G.~Sukthankar (eds.)}  
\acmYear{2020}  
\copyrightyear{2020}  
\acmPrice{}  

\usepackage[utf8]{inputenc} 
\usepackage[T1]{fontenc}    
\usepackage{url}            
\usepackage{booktabs}       
\usepackage{amsfonts}       
\usepackage{nicefrac}       
\usepackage{microtype}      
\usepackage{subfig}
\usepackage{graphicx}
\usepackage{amsmath}
\usepackage{wrapfig}
\usepackage{amssymb}

\usepackage{xcolor}
\hypersetup{
    colorlinks=true,
    linkcolor=blue,
    filecolor=magenta,      
    urlcolor=cyan,
}

\DeclareMathOperator*{\argmax}{argmax}   

\begin{document}

\title{Capacity, Bandwidth, and Compositionality in Emergent Language Learning}
\titlenote{The first two authors contributed equally.}

\author{Cinjon Resnick*} 
\affiliation{\institution{New York University}}
\email{cinjon@nyu.edu} 

\author{Abhinav Gupta*} 
\affiliation{\institution{MILA}}
\email{abhinavg@nyu.edu}

\author{Jakob Foerster}
\affiliation{\institution{Facebook AI Research}}
\email{jnf@fb.com}

\author{Andrew M. Dai}
\affiliation{\institution{Google AI}}
\email{adai@google.com}

\author{Kyunghyun Cho}
\affiliation{\institution{New York University}
\institution{Facebook AI Research}}
\email{kyunghyun.cho@nyu.edu}

\begin{abstract}  
Many recent works have discussed the propensity, or lack thereof, for emergent languages to exhibit properties of natural languages. A favorite in the literature is learning compositionality. We note that most of those works have focused on communicative bandwidth as being of primary importance. While important, it is not the only contributing factor. In this paper, we investigate the learning biases that affect the efficacy and compositionality in multi-agent communication. Our foremost contribution is to explore how the capacity of a neural network impacts its ability to learn a compositional language. We additionally introduce a set of evaluation metrics with which we analyze the learned languages. Our hypothesis is that there should be a specific range of model capacity and channel bandwidth that induces compositional structure in the resulting language and consequently encourages systematic generalization. While we empirically see evidence for the bottom of this range, we curiously do not find evidence for the top part of the range and believe that this is an open question for the community.\footnote{Code is available at \url{https://github.com/backpropper/cbc-emecom}.}
\end{abstract}


\maketitle


\section{Introduction}

Compositional language learning in the context of multi agent emergent communication has been extensively studied \citep{sukhbaatar2016learning, foerster_learning_2016, lazaridou_multi-agent_2016, baroni_linguistic_2019}. These works have found that while most emergent languages do not tend to be compositional, they can be guided towards this attribute through artificial task-specific constraints \citep{kottur2017natural, lee_emergent_2017}. 

In this paper, we focus on how a neural network, specifically a generative one, can learn a compositional language. Moreover, we ask how this can occur without task-specific constraints. To accomplish this, we first define what is a language and what we mean by compositionality. In tandem, we introduce \emph{precision} and \emph{recall}, two metrics that help us measure how well a generative model at large has learned a grammar from a finite set of training instances. We then use a variational autoencoder with a discrete sequence bottleneck to investigate how well the model learns a compositional language, in addition to what affects that learning. This allows us to derive \emph{residual entropy}, a third metric that reliably measures compositionality in our particular environment. We use this metric to cross-validate precision and recall. Finally, while there may be a connection between our study and human-level languages, we do not experiment with human data or languages in this work.

Our environment lets us experiment with a syntactic, compositional language while varying the channel width and the number of parameters, our surrogate for the capacity of a model. Our experiments reveal that our smallest models are only able to solve the task when the channel is wide enough to allow for a surface-level compositional representation. In contrast, large models learn a language as long as the channel is large enough. However, large models also have the ability to memorize the training set. We hypothesize that this memorization would lead to non-compositional representations and overfitting, albeit this does not yet manifest empirically. This setup allows us to test our hypothesis that there is a network capacity above which models will tend to produce languages with non-compositional structure.

\section{Related Work}
There has recently been renewed interest in studies of emergent language \citep{foerster_learning_2016, lazaridou_multi-agent_2016, havrylov_emergence_2017} that originated with works such as \cite{Aibo, steels_1997}. Some of these approaches use referential games \citep{evtimova2018emergent, lazaridou_emergence_2018,lowe*2020on} to produce an emergent language that ideally has properties of human languages, with compositionality being a commonly sought after property \citep{barrett_2018, baroni_linguistic_2019, chaabouni_anti-efficient_2019}.

Our paper is most similar to \cite{kottur2017natural}, which showed that compositional language arose only when certain constraints on the agents are satisfied. While the constraints they examined were either making their models memoryless or having a minimal vocabulary in the language, we hypothesized about the importance for agents to have small capacity relative to the number of disentangled representations (concepts) to which they are exposed. This is more general because both of the scenarios they described fall under the umbrella of reducing model capacity. To ask this, we built a much bigger dataset to illuminate how capacity and channel width effect the resulting compositionality in the language.
Liska\cite{liska_2018} suggests that the average training run for recurrent neural networks does not converge to a compositional solution, but that a large random search will produce compositional solutions. This implies that the optimization approach biases learning, which is also confirmed in our experiments. However, we further analyze other biases. Spike\cite{Spike2017MinimalRF} describes three properties that bias models towards successful learned signaling: the creation and transmission of referential information, a bias against ambiguity, and information loss. This lies on a similar spectrum to our work, but pursues a different intent in that they study biases that lead to optimal signaling; we seek compositionality. Each of \cite{tessa_2016, kirbycompression, zaslavsky_efficient_2018} examine the trade-off between expression and compression in both emergent and natural languages, in addition to how that trade-off affects the learners. We differ in that we target a specific aspect of the agent (capacity) and ask how that aspect biases the learning. Chen\cite{chen_acl} describes how the probability distribution on the set of all strings produced by a recurrent model can be interpreted as a weighted language; this is relevant to our formulation of the language.

\begin{figure}[t]
    \centering
    \includegraphics[width=0.55\linewidth]{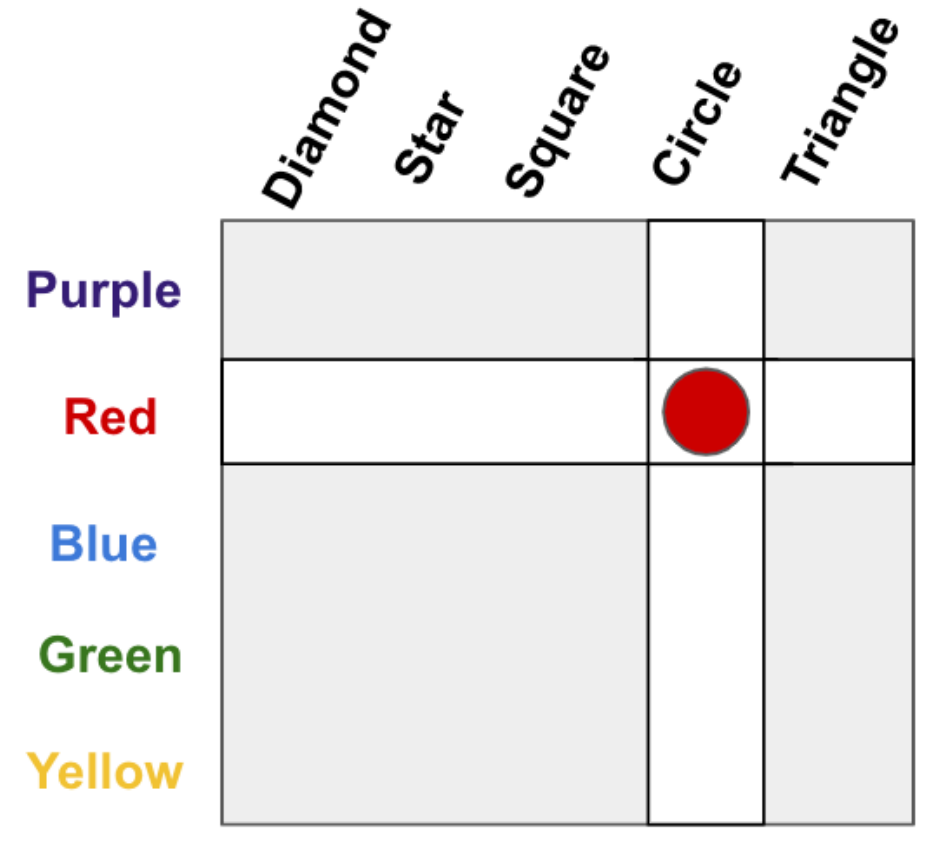}
    \caption{The grid above shows five shapes and five colors. Agents with a non-compositional language can use this shared map to communicate "Red Circle" with only $\lceil\log_2 {5^2}\rceil=5$ bits. If they instead used a compositional language, it would require $\lceil\log_2 5\rceil=3$ bits for each concept for a total of $6$ bits to convey the string. On the other hand, the agent needs $25$ memory slots to store the concepts in the former case but only $10$ slots in the compositional case. This trade-off exemplifies the motivation for our investigation because it suggests that a key driver of compositionality in language is the capacity of an agent relative to the total number of objects in its environment.}
    \label{fig:schematic-diagram}
    \vspace{-3mm}
\end{figure}

Most other works studying compositionality in emergent languages \citep{andreas2018measuring, mordatch_emergence_2017, chaabouni_word-order_2019, lee_countering_2019} have focused on learning interpretable representations. See \cite{hupkes_compositionality_2019} for a broad survey of the different approaches. By and large, these are orthogonal to our work because none pose the question we ask - how does an agent's capacity effect the resulting language's compositionality?

\begin{figure*}[ht]
    \centering
    \subfloat[Precision]{{\includegraphics[width=0.33\linewidth]{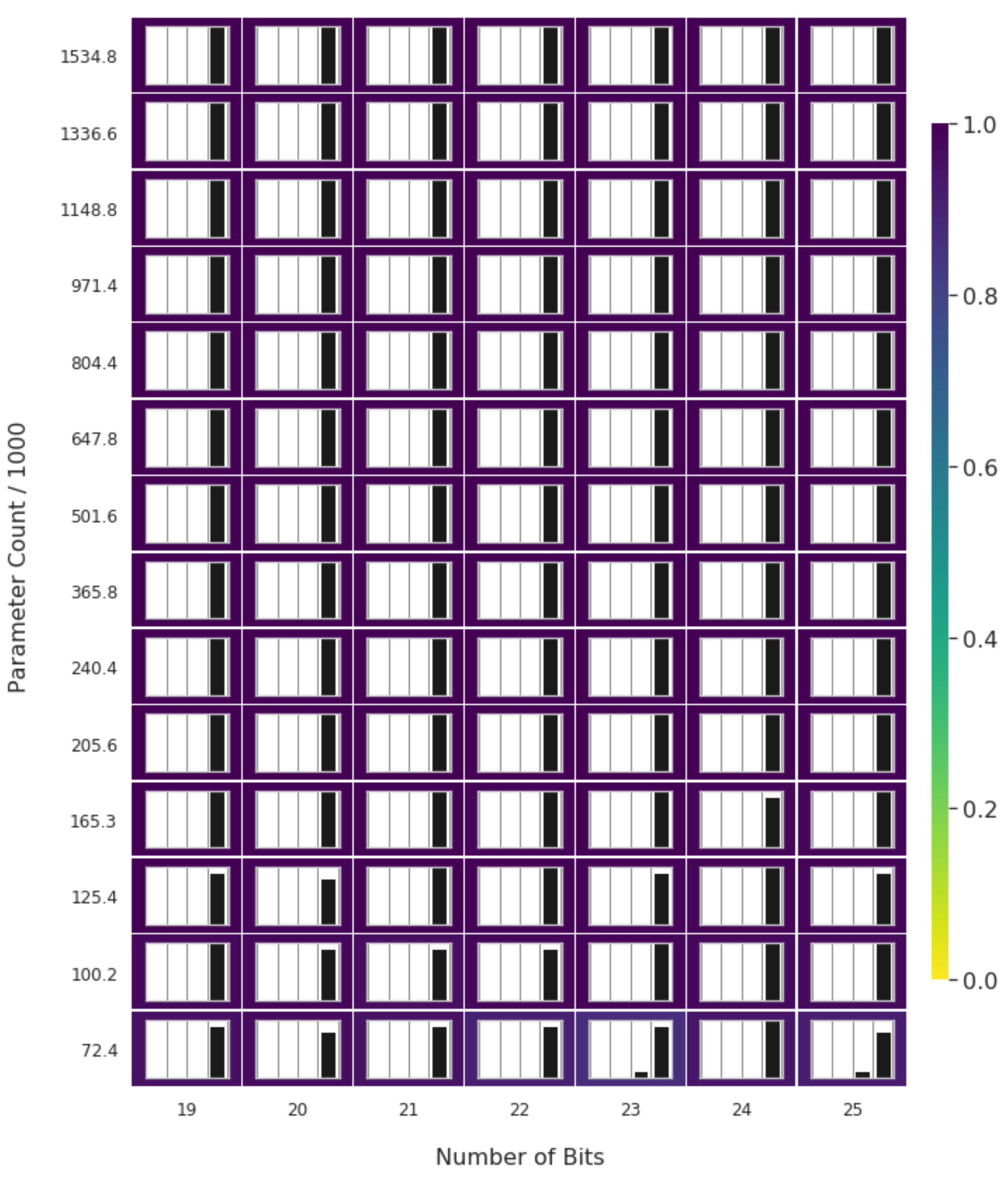}}}
    \hfill
    \subfloat[Recall]{{\includegraphics[width=0.33\linewidth]{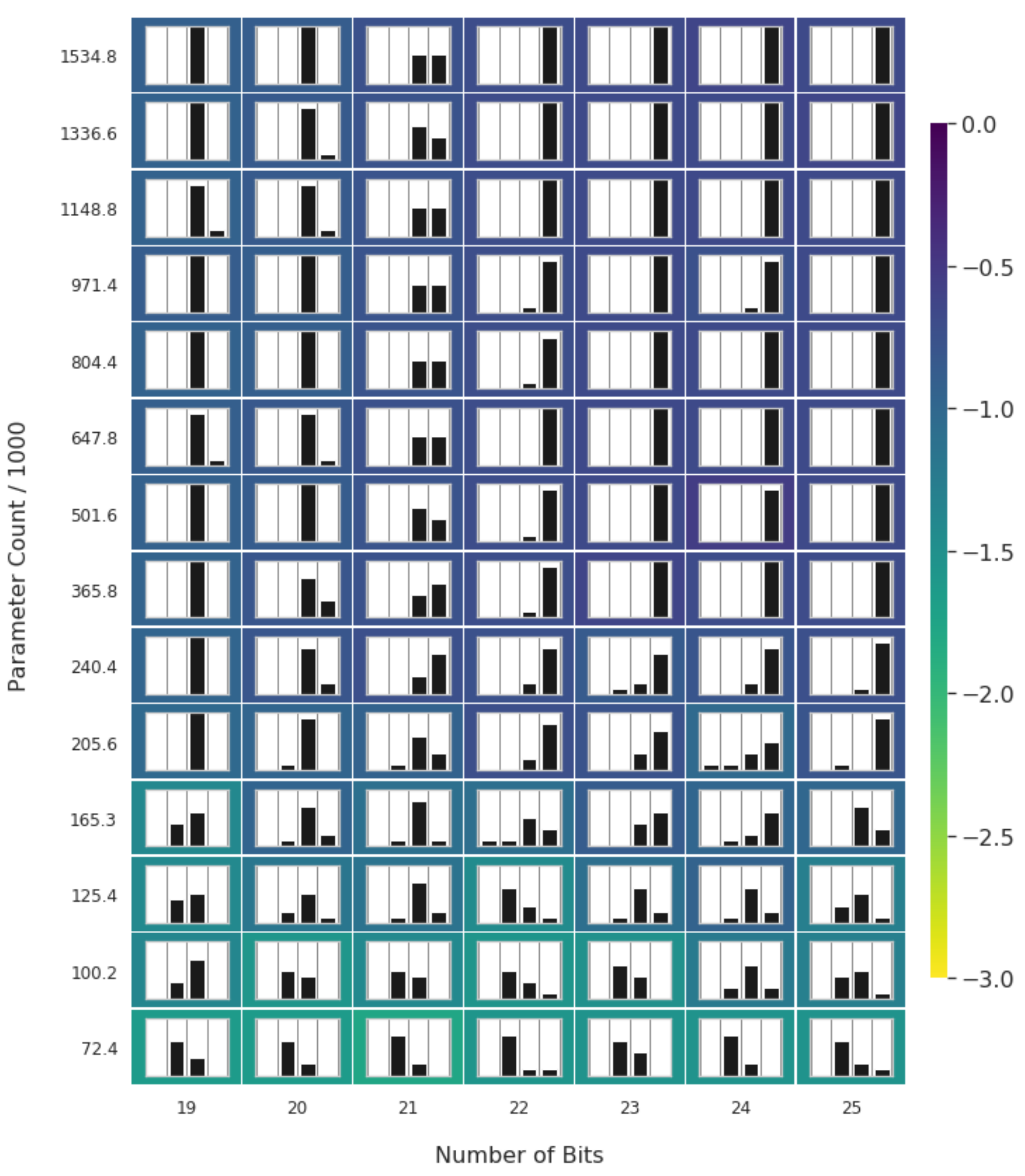}}}
    \hfill
    \subfloat[Entropy]{{\includegraphics[width=0.33\linewidth]{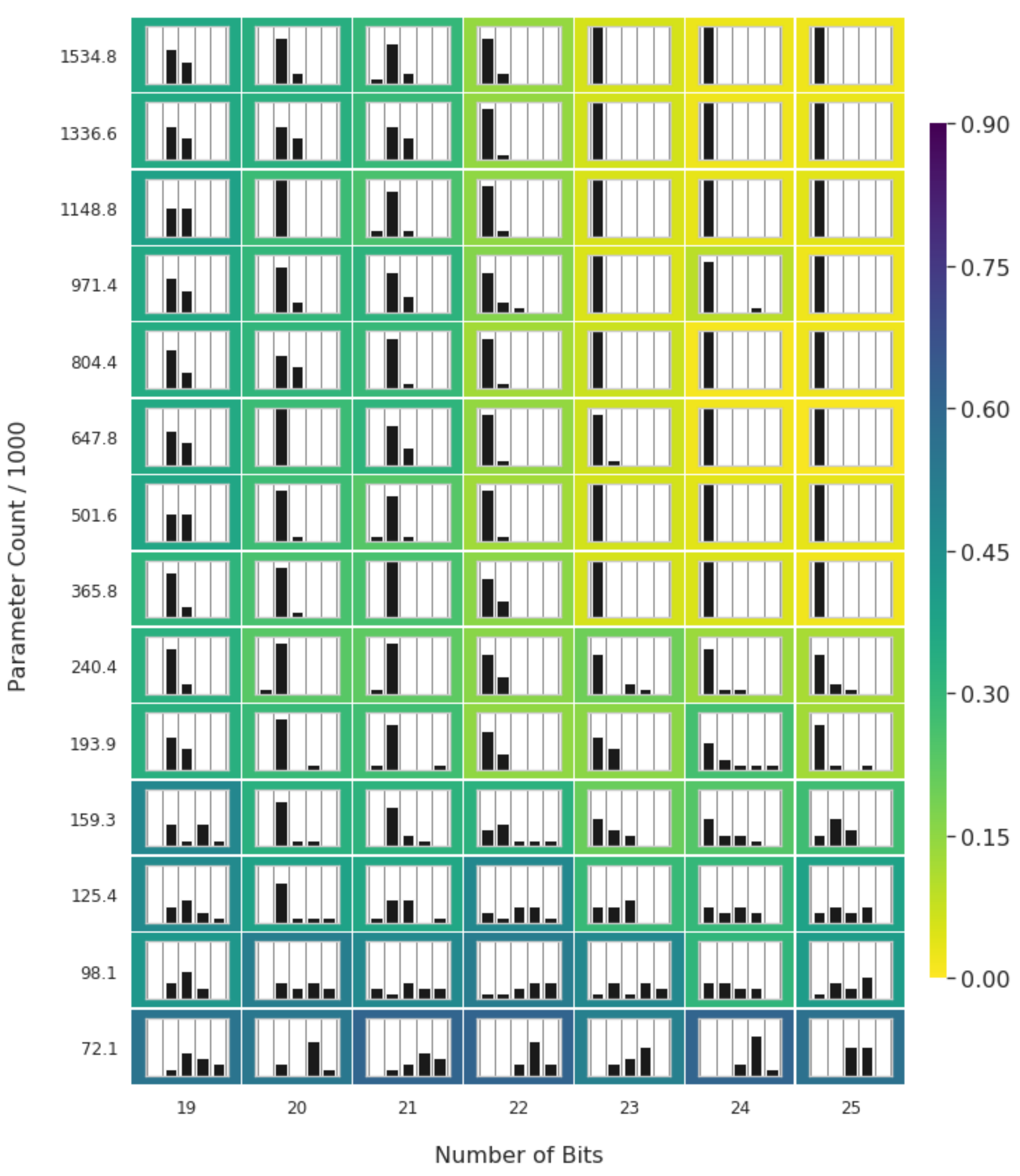}}}
    \caption{Histograms for model A showing precision, recall (defined in \S~\ref{par:eval-metrics}), and entropy (defined in \S~\ref{par:entropymetric}) over the test set. We show results for bits $19$ to $25$ and parameter range $72k$ to $1534k$ (details in \S~\ref{par:model-learning}). Each bit/parameter combination is trained for $10$ seeds over $200k$ steps. 
    Precision and Recall are as described in Eqs.~\eqref{eq:precision-sample} and \eqref{eq:recall-sample} with $M=|D_\text{test}|$ and $N=10000$.
    }
    \label{fig:recall-precision-entropy-A}
\end{figure*}

\section{Compositional Language and Learning}

We start by defining a language and what it means for a language to be compositional. We then discuss what it means for a network to learn a compositional language, based on which we derive evaluation metrics.

\subsection{Compositional Language}
\label{sec:compositional-language}

A {\it language} $L$ is a subset of $\Sigma^*$, where $\Sigma$ denotes an alphabet and $s$ denotes a string: 
\begin{align*}
    \Sigma^* = \left\{ 
    s \mid \forall i=1,\ldots,|s|\quad s_i \in \Sigma \wedge |s| \geq 0 
    \right\}.
\end{align*}
In this paper, we constrain a language to contain only finite-length strings, i.e., $|s| < \infty$, implying that $L$ is a finite language. We use $K_{\max}$ to denote the maximum length of any $s \in L$.

We define a {\it generator} $G$ from which we can sample one valid string $s \in L$ at a time. It never generates an invalid string and generates all the valid strings in $L$ in finite time such that
\begin{align}
\label{eq:generator-language}
s \sim G \Leftrightarrow s \in L.
\end{align}
We define the length $|G|$ of the description of $G$ as the sum of the number of non-terminal symbols $\mathcal{N}$ and the number of production rules $\mathcal{P}$, where $|\mathcal{N}| < \infty$ and $|\mathcal{P}| < \infty$. Each production rule $\rho \in \mathcal{P}$ takes as input an intermediate string $s' \in (\Sigma \cup \mathcal{N})^*$ and outputs another string $s'' \in (\Sigma \cup \mathcal{N})^*$.
The generator starts from an empty string $\varnothing$ and applies an applicable production rule (uniformly selected at random) until the output string consists only of characters from the alphabet (terminal symbols).

\paragraph{Languages and compositionality}
\label{sec:languages}

When the number of such production rules $|\mathcal{P}|$ plus the number of intermediate symbols $|\mathcal{N}|$ is smaller than the size $|L|$ of the language that $G$ generates, we call $L$ a compositional language. In other words, $L$ is compositional if and only if $|L| > |\mathcal{P}|+|\mathcal{N}|$.

One such example is when we have sixty characters in the alphabet, $\Sigma=\left\{ 0, 1, 2, \ldots, 59 \right\}$, and six intermediate symbols, $\mathcal{N}=\left\{ C_1, C_2, \ldots, C_6\right\}$, for a total of $6 \times 10 + 1$ production rules $\mathcal{P}$:
\begin{itemize}
    \item $\varnothing \to C_1 C_2 C_3 C_4 C_5 C_6$.
    \item For each $i \in [1, 6]$, $C_i \to w$, where $w \in \left\{10i-10, \ldots, 10i - 1\right\}$
\end{itemize}
From these production rules and intermediate symbols, we obtain a language of size $10^{6} \gg 67 = |\mathcal{P}|+|\mathcal{N}|$. We thus consider this language to be compositional and will use it in our experiments.

\subsection{Learning a language}

We consider the problem of learning an underlying language $L^\star$ from  a finite set of training strings randomly drawn from it:
\begin{align*}
D = \left\{ 
s | s \sim G^\star
\right\}
\end{align*}
where $G^\star$ is the minimal length generator associated with $L^\star$. We assume $|D| \ll |L^\star|$ and our goal is to use $D$ to learn a language $L$ that approximates $L^\star$ as well as possible. We know that there exists an equivalent generator $G$ for $L$, and so our problem becomes estimating a generator from this finite set rather than reconstructing an entire set of strings belonging to the original language $L^*$. 

We cast the problem of estimating a generator $G$ as density modeling, in which case the goal is to estimate a distribution $p(s)$. Sampling from $p(s)$ is equivalent to generating a string from the generator $G$. Language learning is then 
\begin{align}
    \label{eq:language-learning-generic}
    \max_{p \in \mathcal{M}} \frac{1}{|D|} \sum_{n=1}^{|D|} \log p(s_n) + \lambda R(p),
\end{align}
where $R$ is a regularization term, $\lambda$ its strength, and $\mathcal{M}$ is a model space. 

\paragraph{Evaluation metrics}
\label{par:eval-metrics}
When the language was learned perfectly, any string sampled from the learned distribution $p(s)$ must belong to $L^\star$. Also, any string in $L^\star$ must be assigned a non-zero probability under $p(s)$. Otherwise, the set of strings generated from this generator, implicitly defined via $p(s)$, is not identical to the original language $L^\star$. This observation leads to two metrics for evaluating the quality of the estimated language with the distribution $p(s)$, \textit{precision} and \textit{recall}:
\begin{align}
    &\text{Precision}(L^\star, p) = \frac{1}{|L^\star|} \sum_{s \in L} \mathbb{I}(s \in L^\star)
    \\
    &\text{Recall}(L^\star, p) =  \sum_{s \in 
L^\star} \log p(s)
\end{align}
where $\mathbb{I}(x)$ is the indicator function. These metrics are designed to be fit for any compositional structure rather than one-off evaluation approaches. Because these are often intractable to compute, we approximate them using Monte-Carlo by sampling $N$ samples from $p(s)$ for calculating precision and $M$ uniform samples from $L^\star$ for calculating recall.
\begin{align}
\label{eq:precision-sample}
    &\text{Precision}(L^\star, p) \approx \frac{1}{N} \sum_{n=1}^N \mathbb{I}(s_n \in L^\star) 
    \\
\label{eq:recall-sample}
    &\text{Recall}(L^\star, p) \approx \sum_{m=1}^M 
    \log p(s_m),
\end{align}
where $s_n \sim p(s)$ and $s_m$ is a uniform sample from $L^\star$. 

\subsection{Compositionality, learning, and capacity}
\label{sec:compositionality-generic}

When learning an underlying compositional language $L^\star$, there are three possible outcomes:

{\bf Overfitting:} $p(s)$ could memorize all the strings that were presented when solving Eq.~\eqref{eq:language-learning-generic} and assign non-zero probabilities to those strings and zero probabilities to all others. This would maximize precision, but recall will be low as the estimated generator does not cover $L^\star$. 

{\bf Systematic generalization:} $p(s)$ could capture the underlying compositional structures of $L^\star$ characterized by the production rules $\mathcal{P}$ and intermediate symbols $\mathcal{N}$. In this case, $p(s)$ will assign non-zero probabilities to all the strings that are reachable via these production rules (and zero probability to all others) and generalize to strings from $L^\star$ that were unseen during training, leading to high precision and recall. This behavior was characterized in \citet{lake_compgen}. 

{\bf Failure:} $p(s)$ may neither memorize the entire training set nor capture the underlying production rules and intermediate symbols, resulting in both low precision and recall. 

We hypothesize that a major factor determining the compositionality of the resulting language is the capacity of the most complicated distribution $p(s)$ within the model space $\mathcal{M}$.\footnote{
    As the definition of a model's capacity heavily depends on the specific construction, we do not concretely define it here but do later when we introduce a specific family of models with which we run experiments. 
}
When the model capacity is too high, the first case of total memorization is likely. When it is too low, the third case of catastrophic failure will happen. Only when the model capacity is just right will language learning correctly capture the compositional structure underlying the original language $L^\star$ and exhibit systematic generalization \citep{bahdanau2018systematic}. 
We empirically investigate this hypothesis using a neural network as a language learner, in particular a variational autoencoder with a discrete sequence bottleneck. This admits the interpretation of a two-player ReferIt game~\citep{lewis1969convention} in addition to being a density estimator of $p(s)$. Together, these let us use recall and precision to analyze the resulting emergent language.

\section{Variational autoencoders and their capacity}

A variational autoencoder \citep{kingma2014auto} consists of two neural networks which are often referred to as an encoder $f_{\theta}$, a decoder $g_{\phi}$, and a prior $p_{\lambda}$. These two networks are jointly updated to maximize the variational lower bound to the marginal log-probability of training instances $s$:
\begin{align}
    \label{eq:variational-lowerbound}
    \mathcal{L}(\theta, \phi, \lambda; s) = \mathbb{E}_{z \sim f_{\theta}(z|s)}\left[ \log g_{\phi}(s | z) \right]
    - \text{KL}(f_{\theta}(z|s) \| p_{\lambda} ).
\end{align}

We use $\mathcal{L}$ as a proxy to the true $p(s)$ captured by this model. Once trained, we can efficiently sample $\tilde{z}$ from the prior $p_{\lambda}$ and then sample a string from $g_{\phi}(s | \tilde{z})$.

The usual formulation of variational autoencoders uses a continuous latent variable $z$, which conveniently admits reparametrization that reduces the variance in the gradient estimate. However, this infinitely large latent variable space makes it difficult to understand the resulting capacity of the model. We thus constrain the latent variable to be a binary string of a fixed length $l$, i.e., $z \in \left\{ 0, 1 \right\}^l$. Assuming deterministic decoding, i.e., $\argmax_s \log g_\phi(s | z)$, this puts a strict upperbound of $2^l$ on the size of the language $|L|$ captured by the variational autoencoder. 

\subsection{Variational autoencoder as a communication channel}
\label{sec:variationalauto}

As described above, using variational autoencoders with a discrete sequence bottleneck allows us to analyze the capacity of the model in terms of computation and bandwidth. We can now interpret this variational autoencoder as a communication channel in which a novel protocol must emerge as a by-product of learning. We will refer to the encoder as the speaker and the decoder as the listener when deployed in this communication game. If each string $s \in L^\star$ in the original language is a description of underlying concepts, then the goal of the speaker $f_\theta$ is to encode those concepts in a binary string $z$ following an emergent communication protocol. The listener $g_\phi$ receives this string and must interpret which set of concepts were originally seen by the speaker. 

\begin{figure}[ht]
    \centering
    \subfloat[Model A Training]{{\includegraphics[width=0.48\linewidth]{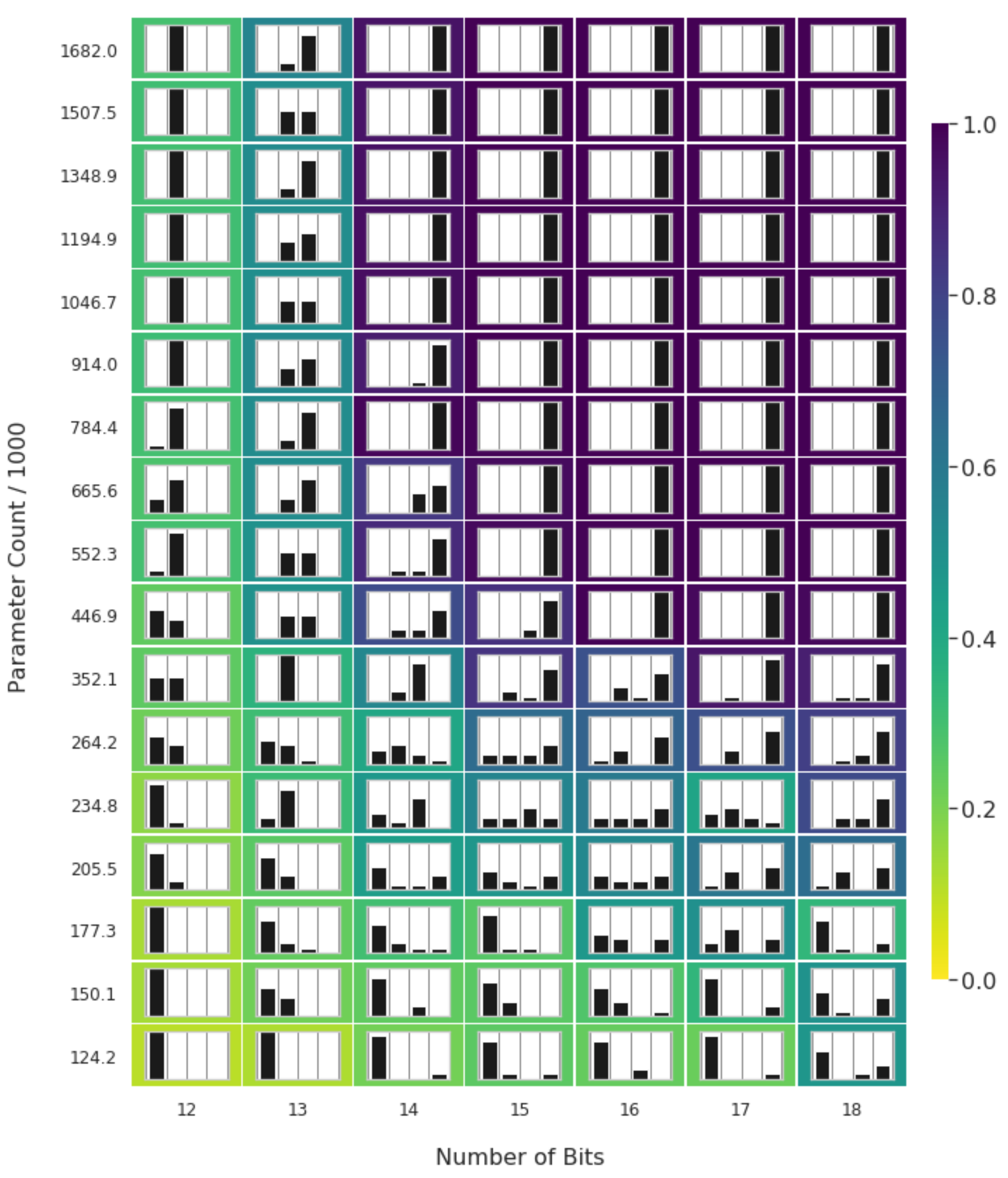}}}
    \hfill
    \subfloat[Model A Testing]{{\includegraphics[width=0.48\linewidth]{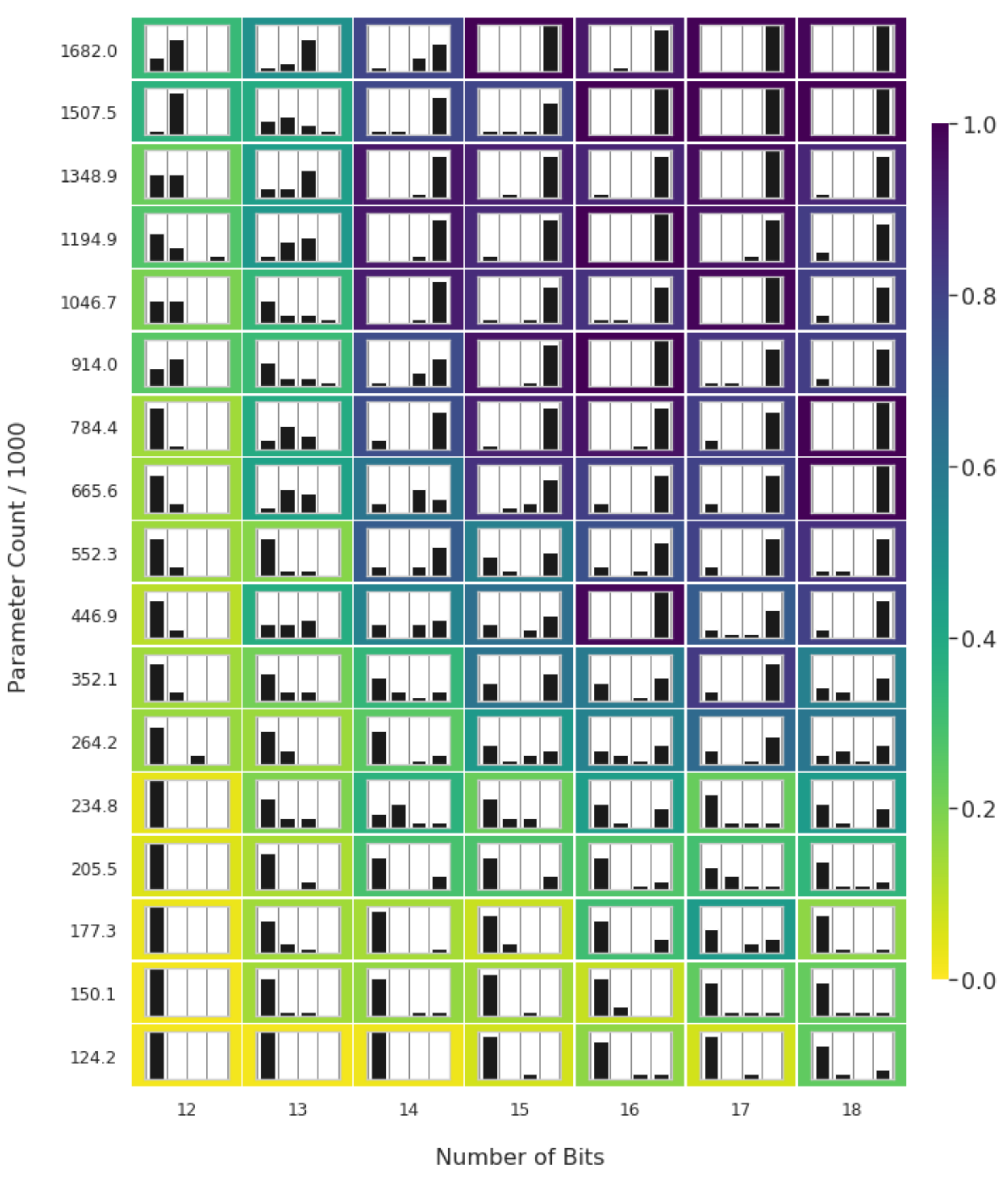}}}
    \hfill
    \subfloat[Model A Recall]{{\includegraphics[width=0.48\linewidth]{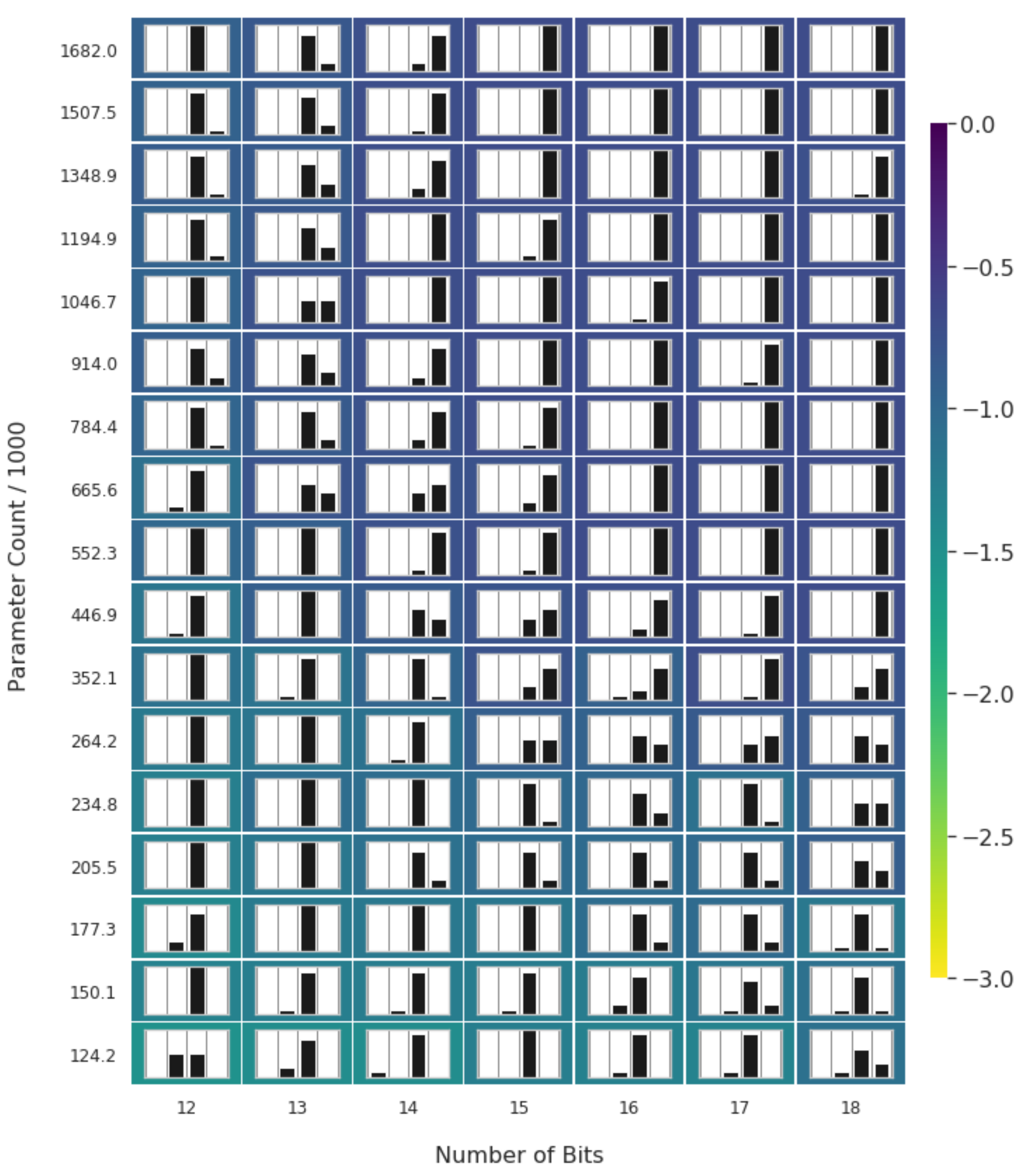}}}
    \hfill
    \subfloat[Model A Entropy]{{\includegraphics[width=0.48\linewidth]{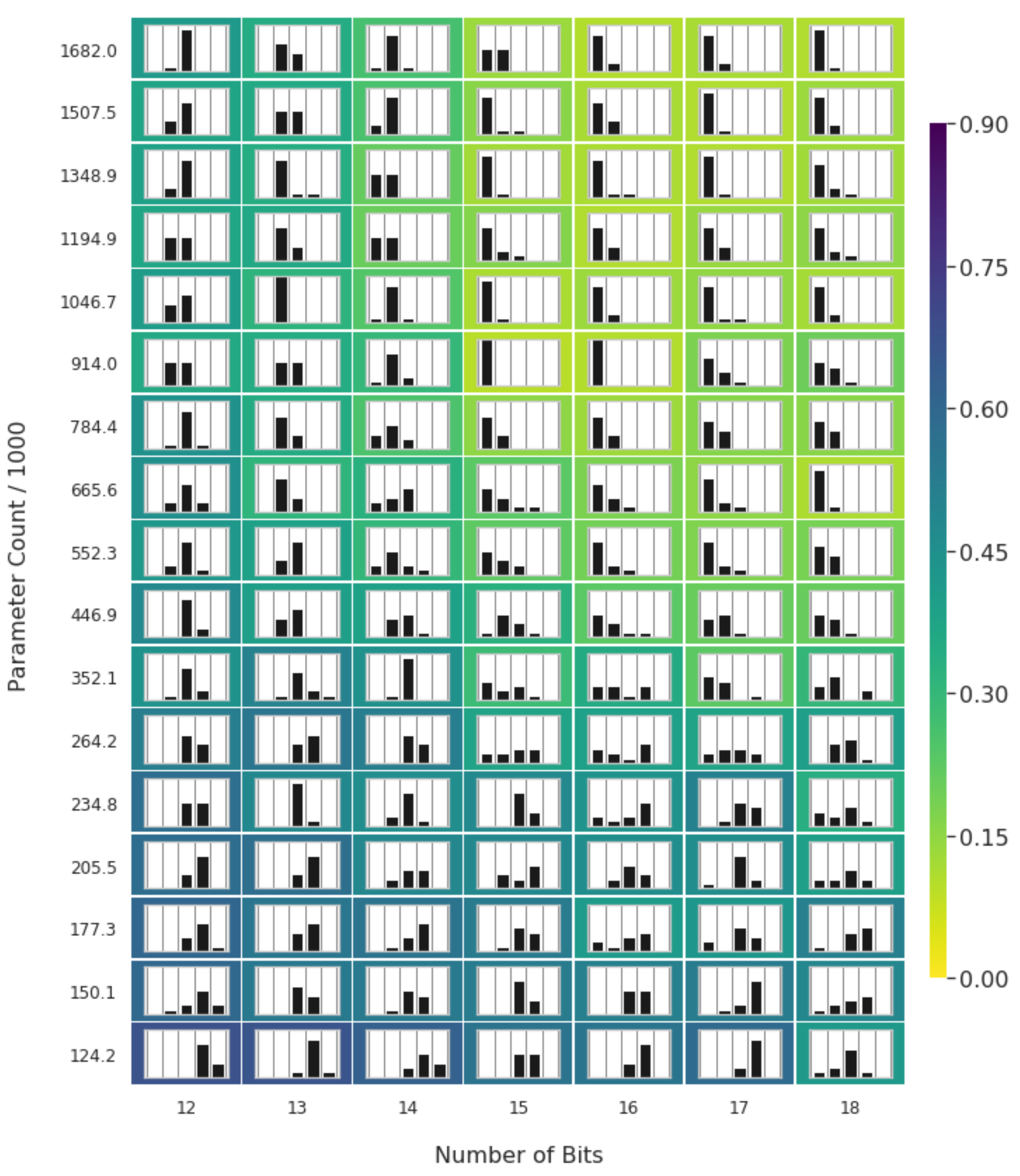}}}
    \caption{Results when running Model A with $N=4$ categories instead of $N=6$. We show results for bits $12$ to $18$ and parameter range $124k$ to $1682k$. We need at least $\lceil \log_2 10^4 \rceil = 14$ bits to cover all the input combinations. Observe that there is not much difference to the $N=6$ scenario show in fig~\ref{fig:recall-precision-entropy-A}. See \S~\ref{par:eval-metrics} and \S\ref{par:entropymetric} for the definitions of recall and entropy.}
    \label{fig:small-model}
\end{figure}

\begin{figure*}[ht]
    \centering
    \includegraphics[width=1.0\linewidth, height=5cm]{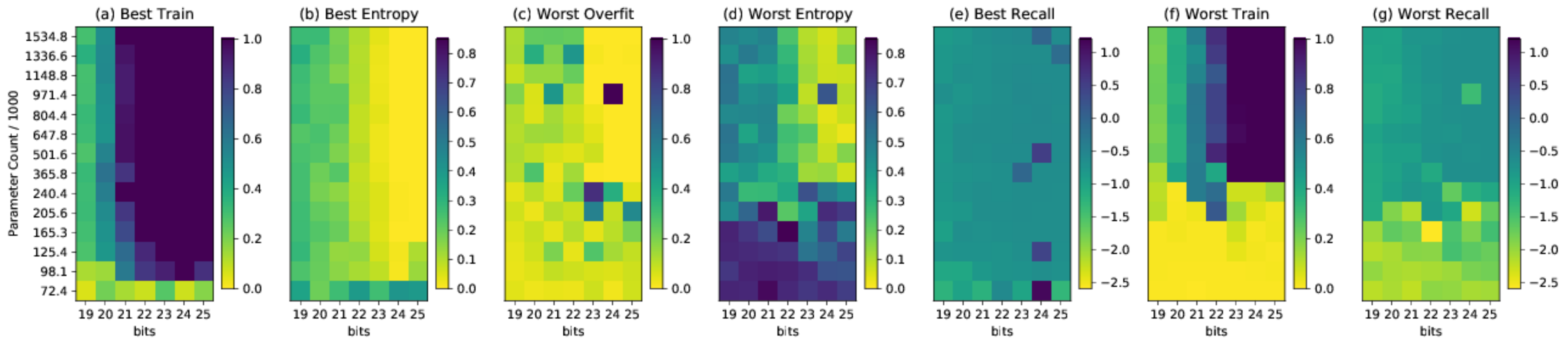}
        \caption{Main results for model A showing best and worst performances of the proposed metrics over $10$ seeds. See Section \ref{sec:results} for detailed analysis. Panels (a) and (f) show the accuracy of the training data, (b) and (d) show entropy, (e) and (g) show recall over the test data, and (c) plots the max difference in accuracy between training and test.} 
    \label{fig:master_plot}
\end{figure*}

\paragraph{Our setup}

We simplify and assume that each of the characters in the string $s \in L^\star$ correspond to underlying concepts. While the inputs are ordered according to the sequential concepts, our model encodes them using a bag of words (BoW) representation.

The speaker $f_\theta$ is parameterized using a recurrent policy which
receives the sequence of concatenated one-hot input tokens of $s$ and converts each of them to an embedding. It then runs an LSTM \citep{Hochreiter:1997} non-autoregressively for $l$ timesteps taking the flattened representation of the input embeddings as its input and linearly projecting each result to a probability distribution over $\{0, 1\}$. This results in a sequential Bernoulli distribution over $l$ latent variables:
$$f_{\theta}(z | s) = \prod_{t=1}^{l} p(z_t | s;\theta)$$
From this distribution, we can sample a latent string $z=(z_1, \ldots, z_l)$. 

The listener $g_\phi$ receives $z$ and uses a BoW representation
to encode them into its own embedding space. Taking the flattened representation of these embeddings as input, we run an LSTM for $|\mathcal{N}|$ time steps, each time outputting a probability distribution over the full alphabet $\Sigma$: 
$$ g_{\phi}(s|z) = \prod_{j=1}^{|\mathcal{N}|} p(s_j|z;\phi) $$
    
To train the whole system end-to-end \citep{sukhbaatar2016learning, mordatch_emergence_2017} via backpropogation, we apply a continuous approximation to $z_t$ that depends on a learned temperature parameter $\tau$. We use the `straight-through` version of Gumbel-Softmax \citep{jang_categorical_2016, maddison2016concrete} to convert the continuous distribution to a discrete distribution for each $z_t$. This corresponds to the original discrete distribution in the zero-temperature limit. The final sequence of one hot vectors encoding $z$ is our \textit{message}, which is passed to the listener $g_\phi$. If $G_j \sim \textrm{Gumbel}(0, 1)$ and the Bernoulli random variable corresponding to $z_t$ has class probabilities $z_{t_0}$ and $z_{t_1}$, then $ z_t = \texttt{one\_hot}(\underset{j}{\arg\max} [G_j + \log z_{t_j}]) $.

The prior $p_{\lambda}$ encodes the \textit{message} $z$ using a BoW representation. It gives the probability of $z$ according to the prior (binary) distribution for each $z_t$ and is defined as:
$$ p_{\lambda}(z) = \prod_{t=1}^{l}p(z_t|\lambda).$$

This can be used both to compute the prior probability of a latent string and also to efficiently sample from $p_{\lambda}$ using ancestral sampling. Penalizing the KL divergence between the speaker's distribution and the prior distribution in Eq.~\eqref{eq:variational-lowerbound} encourages the emergent protocol to use latent strings that are as diverse as possible.

\subsection{Capacity of a variational autoencoder with discrete sequence bottleneck}
\label{sec:capacity}

This view of a variational autoencoder with discrete sequence bottleneck presents an opportunity for us to separate the model's capacity into two parts. The first part is the capacity of the communication channel, imposed by the size of the latent variable. As described earlier, the size of the original language $L^\star$ that can be perfectly captured by this model is strictly upper bounded by $2^l$, where $l$ is the preset length of the latent string $z$. If $l < \log_2 |L^\star|$, the model will not be able to learn the language completely, although it may memorize all the training strings $s \in D$ if $l \geq \log_2 |D|$. A resulting question is whether $2^l \geq |L^\star|$ is a sufficient condition for the model to learn $L^\star$ from a finite set of training strings.

The second part involves the capacity of the speaker and listener to map between the latent variable $z$ and a string $s$ in the original language $L^\star$. Taking the parameter count as a proxy to the number of patterns that could be memorized by a neural network,\footnote{
This is true in certain scenarios such as radial-basis function networks.
}
we can argue that the problem can be solved if the speaker and listener each have $\Omega(l|L^\star|)$ parameters, in which case they can implement a hashmap between a string in the original language $L^\star$ and that of the learned latent language defined by the $z$ strings.

However, when the underlying language is compositional as defined in \S\ref{sec:compositional-language}, we can have a much more compact representation of the entire language than a hashmap. Given the status quo understanding of neural networks, it is impossible to correlate the parameter count with the language specification (production rules and intermediate symbols) and the complexity of using that language. It is however reasonable to assume that there is a monotonic relationship between the number of parameters $|\theta|$, or $|\phi|$, and the capacity of the network to encode the compositional structures underlying the original language \citep{Collins_2017}. Thus, we use the parameter count as a proxy to measure the capacity.

In summary, there are two axes in determining the capacity of the proposed model: the length of the latent sequence $l$ and the number of parameters $|\theta|$ ($|\phi|$) in the speaker (listener).\footnote{
    We design the variational autoencoder to be symmetric so that the parameter counts of the speaker and the listener are roughly the same.
}
We vary these two quantities in the experiments and investigate how they affect compositional language learning by the model.

\subsection{Implications and hypotheses on compositionality}

Under this framework for language learning, we can make the following observations:
\begin{itemize}
\item If the length of the latent sequence $l < \log_2 | L^\star |$, it is impossible for the model to avoid the failure case because there will be $| L^\star | - 2^l$ strings in $L^\star$ that cannot be generated from the trained model. Consequently, recall cannot be maximized. However, this may be difficult to check using the sample-based estimate as the chance of sampling $s \in L^\star \backslash \int g_{\phi}(s | z)p_\lambda(z) \text{d}z$ decreases proportionally to the size of $L^\star$. This is especially true when the gap $| L^\star | - 2^l$ is narrow.

\item When $l \geq \log_2 | L^\star |$, there are three cases. The first is when there are not enough parameters $\theta$ to learn the underlying compositional grammar given by $\mathcal{P}$, $\mathcal{N}$, and $\Sigma$, in which case $L^\star$ cannot be learned. The second case is when the number of parameters $|\theta|$ is greater than that required to store all the training strings, i.e., $|\theta| = O(l|D|)$. Here, it is highly likely for the model to overfit as it can map each training string with a unique latent string without having to learn any of $L^\star$'s compositional structure. Lastly, when the number of parameters lies in between these two poles, we hypothesize that the model will capture the underlying compositional structure and exhibit systematic generalization.
\end{itemize}
In short, we hypothesize that the effectiveness of compositional language learning is maximized when both the length of the latent sequence is large enough ($l \geq \log |L^\star|$), and the number of parameters $|\theta|$ is between $k(|\mathcal{P}| + |\mathcal{N}| + |\Sigma|)$ and $kl|D|$ for some positive integer $k$. Our experiments test this by varying the length of the latent sequence $l$ and the number of parameters $|\theta|$ while checking the sample-based estimates of precision and recall (Eq.~\eqref{eq:precision-sample}--\eqref{eq:recall-sample}). 

\section{Experiments}

\paragraph{Data}

\begin{table}[h]
\centering
{
\small
\begin{tabular}{c|c|c}
& Model A & Model B \\
\toprule
speaker Total & 708k & 670k \\
listener Total & 825k & 690k \\
speaker Embedding & 100 & 40 \\
speaker LSTM & 200 & 300 \\
speaker Linear & 300 & 60 \\
listener Embedding & 300 & 125 \\
listener LSTM & 300 & 325 \\
\end{tabular}
}
\caption{The base hyperparameter counts used in our experiments for each of models A and B.}
\label{table:model-hps}
\end{table}

As described in \S\ref{sec:languages}, we run experiments where the size of the language is much larger than the number of production rules. The task is to communicate $6$ concepts, each of which have $10$ possible values with a total dataset size of $10^6$. We build three finite datasets $D_{\text{train}}$, $D_{\text{val}}$, $D_{\text{test}}$:
\begin{align*}
    &S_{\text{train}} = \left\{ 
    s \in L^\star \mid
    s \neq (*C^{\text{val}}_1*C^{\text{val}}_2*)
    \wedge
    s \neq (*C^{\text{test}}_1*C^{\text{test}}_2*)
    \right\}\\
    &S_{\text{val}} = \left\{ 
    s = (*C^{\text{val}}_1*C^{\text{val}}_2*)
    \right\} \\
    &S_{\text{test}} = \left\{ 
    s = (*C^{\text{test}}_1*C^{\text{test}}_2*)
    \wedge
    s \notin D_{\text{val}}
    \right\} \\
    &D_{\text{train}} = \text{subsample}\left(S_{\text{train}}, N_{\text{train}} \right)
    \\
    &D_{\text{val}} = \text{subsample}\left(S_{\text{val}}, N_{\text{val}} \right)
    \\
    &D_{\text{test}} = \text{subsample}\left(S_{\text{test}}, N_{\text{val}} \right),
\end{align*}
where $\text{subsample}(S, N)$ uniformly selects $N$ random items from $S$ without replacement. The randomly selected concept values $(C^{\text{val}}_1,C^{\text{val}}_2)$ and  $(C^{\text{test}}_1,C^{\text{test}}_2)$ ensure that concept combinations are unique to each set. The $*$ symbol refers to any number of concepts, as in regular expressions. 

\paragraph{Models and Learning}
\label{par:model-learning}
We train the proposed variational autoencoder (described in \S\ref{sec:variationalauto}) on $D_{\text{train}}$, using the Adam optimizer \citep{Kingma2015AdamAM} with a learning rate of $3 \times 10^{-3}$, weight decay coefficient of $10^{-4}$ and a batch size of $1000$. The Gumbel-Softmax temperature parameter $\tau$ in is initialized to 1. Since systematic generalization may only happen in some training runs \citep{weber_2018}, each model is trained for each of $10$ seeds over $200k$ steps. We trained our models using \citep{pytorch_NIPS2019}.

We train two sets of models. Each set is built from an independent base model, the architectures of which are described in Table~\ref{table:model-hps}. We gradually decrease the number of LSTM units from the base model by a factor $\alpha \in \left(0, 1\right]$. This is how we control the number of parameters ($|\theta|$ and $|\phi|$), a factor we hypothesize to influence the resulting compositionality. We obtain seven models from each of these by varying the length of the latent sequence $l$ from $\left\{ 19, 20, 21, 22, 23, 24, 25\right\}$. These were chosen because we both wanted to show a range of bits and because we need at least $20$ bits to cover the $10^6$ strings in $L^*$ ($\lceil \log_2 10^6 \rceil = 20$).

\begin{figure}[h]
    \centering
    \subfloat[Model B Precision]{{\includegraphics[width=0.48\linewidth]{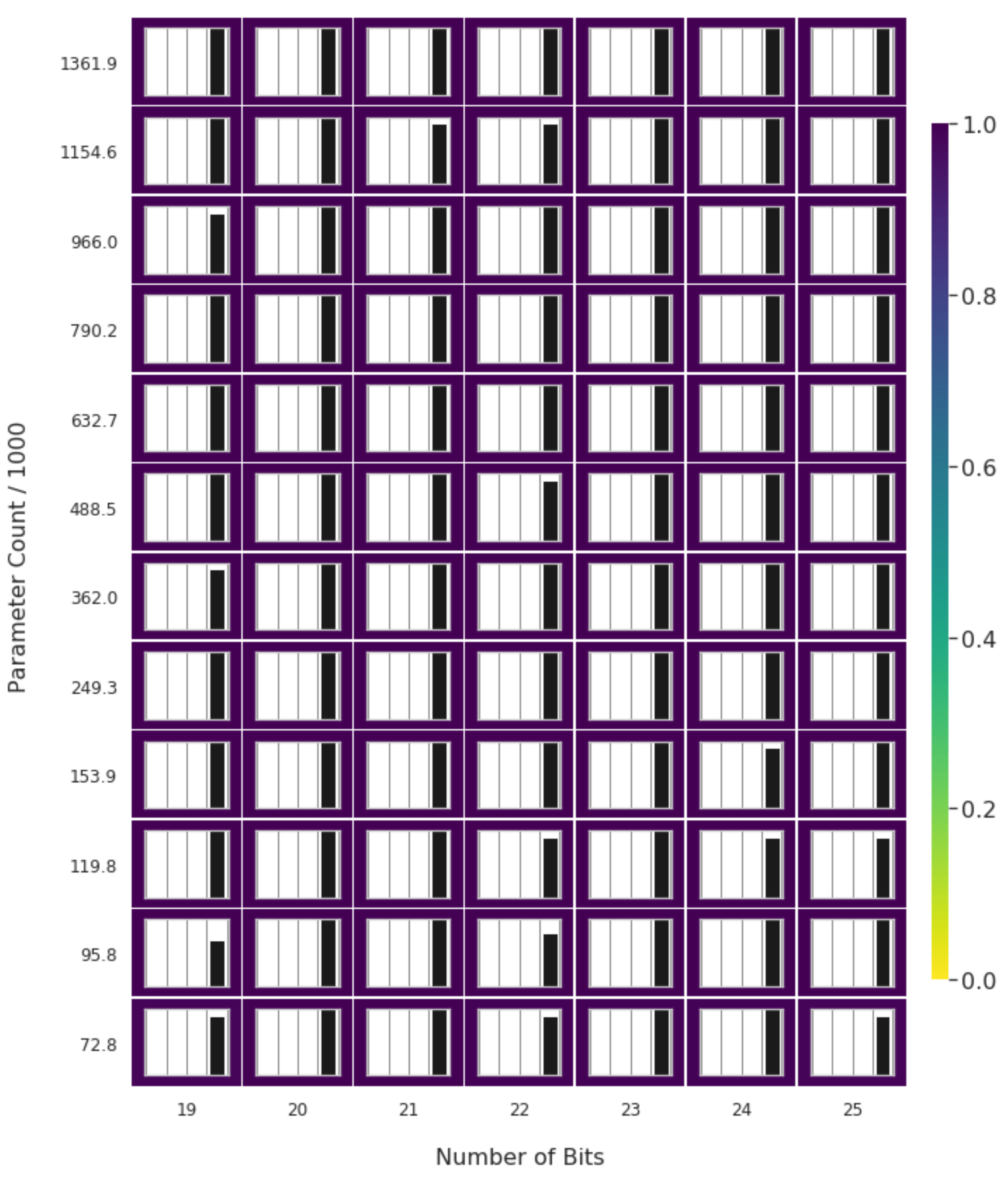}}}
    \hfill
    \subfloat[Model B Recall]{{\includegraphics[width=0.48\linewidth]{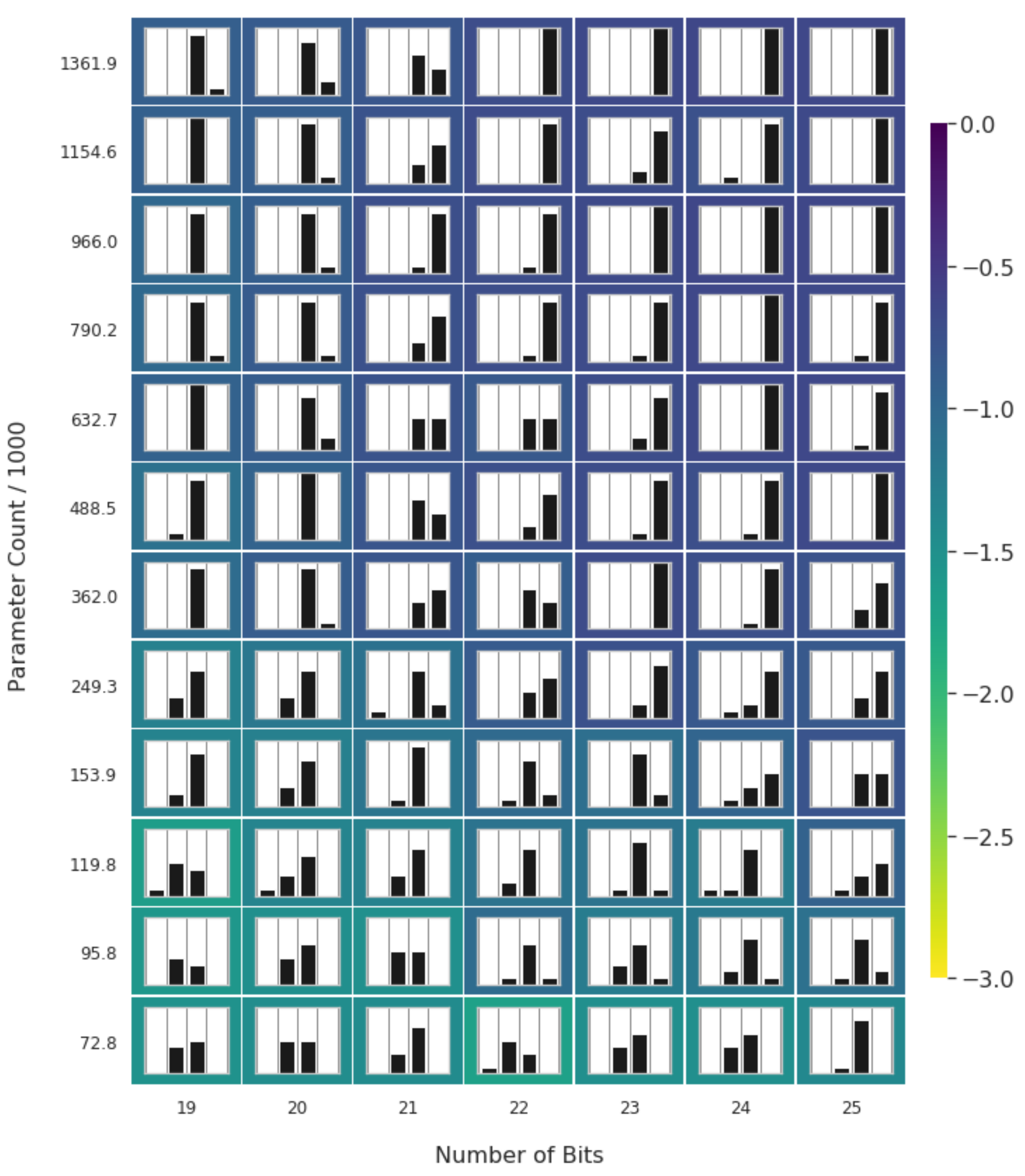}}}
    \qquad    
    \caption{Precision and recall for model B. Like model A, model B has perfect precision. However, its recall chart shows a different story. The first takeaway is that while there is still a strong region in the top right bounded below by $\sim 360k$ parameters, it does not extend to $22$ bits on the left side. This supports our notion of a minimal capacity threshold but adds a wrinkle in that this architecture influences the model's ability to succeed with fewer bits.}
    \label{fig:recall-precision-B}
\end{figure}

\begin{figure}[h]
    \centering
    \includegraphics[width=0.67\linewidth]{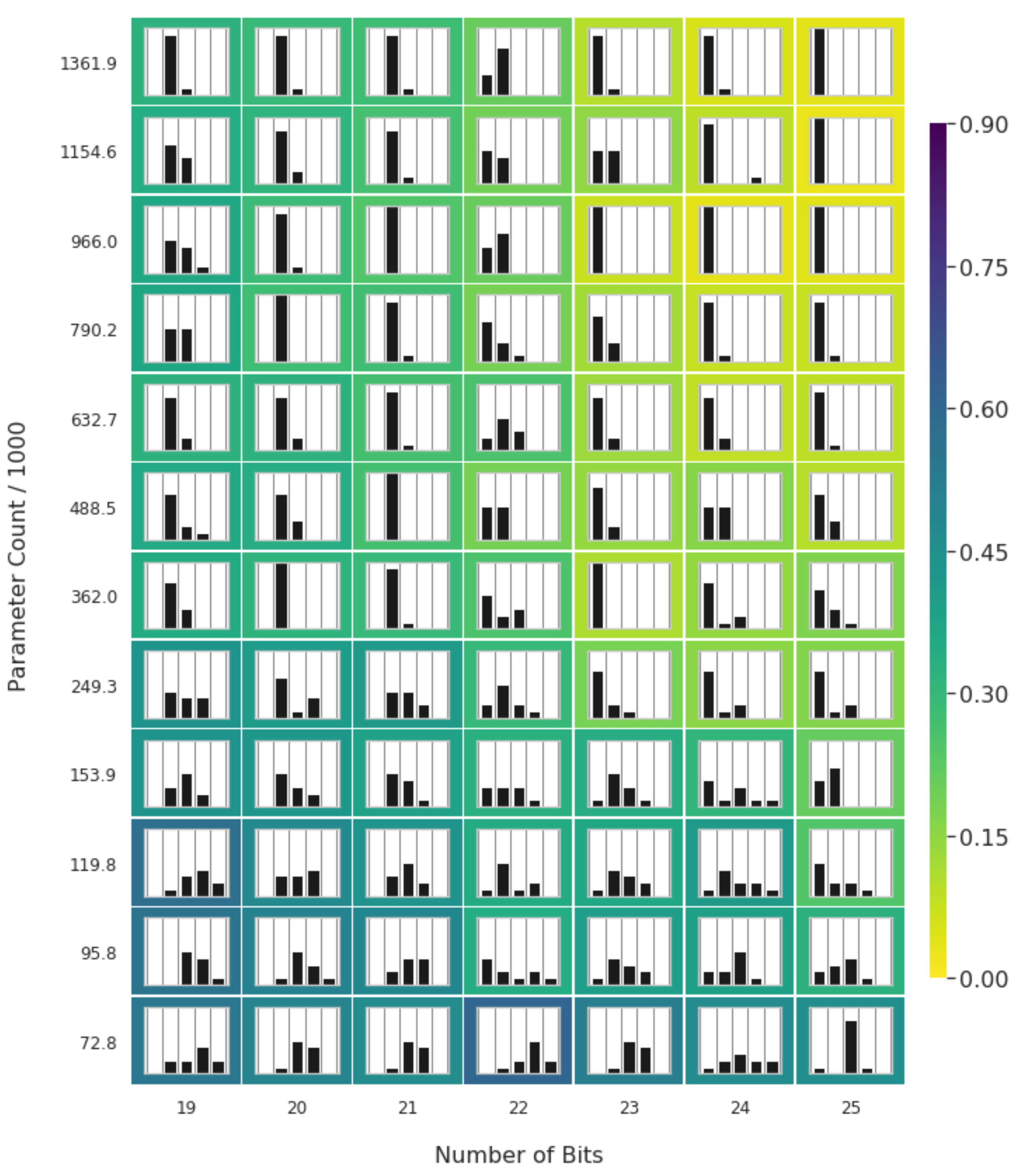}
    \caption{Entropy metric for model B as described in \S\ref{par:entropymetric}. Similar to model A, we see that model B's entropy supports the view we had of its recall in Figure \ref{fig:recall-precision-B} - there is a strong region in the top right that supports the notion of a minimal capacity and bit threshold.}
    \label{fig:entropymetric-B}
\end{figure}

Note that results for the two models are similar and so we arbitrarily spotlight one of them - model A. We additionally show the results for model B in Figs. \ref{fig:recall-precision-B}, \ref{fig:entropymetric-B}, and \ref{fig:efficacy}, but do not dive into its results in the main section.

\subsection{Evaluation: Residual Entropy}
\label{par:entropymetric}

Our setup allows us to design a metric by which we can check the compositionality of the learned language $L$ by examining how the underlying concepts are described by a string. For instance, $(2,11,24,31,44,56) \in L^\star$ describes that $C_1=2$, $C_2=11$, $C_3=24$, $C_4=31$, $C_5=44$ and $C_6=56$. Furthermore, we know that the value of a concept $C_i$ is independent of the other concepts $C_{j\neq i}$, and so our custom generative setup with a discrete latent sequence allows us to inspect a learned language $L$ by considering $z$.

Let $p$ be a sequence of partitions of $\left\{1, 2, \ldots, l\right\}$. We define the degree of compositionality as the ratio between the variability of each concept $C_i$ and the variability explained by a latent subsequence $z[p_i]$ indexed by an associated partition $p_i$. More formally, the degree of compositionality given the partition sequence $p$ is defined as a residual entropy
\begin{align*}
    \text{re}(p, L, L^\star) = \frac{1}{|\mathcal{N}|} \sum_{i=1}^{|\mathcal{N}|} \mathcal{H}_{L}(C_i|z[p_i])/\mathcal{H}_{L^\star}(C_i)
\end{align*}
where there are $|\mathcal{N}|$ concepts by the definition of our language. When each term inside the summation is close to zero, it implies that a subsequence $z[p_i]$ explains most of the variability of the specific concept $C_i$, and we consider this situation compositional. The residual entropy of a trained model is then the smallest $\text{re}(p)$ over all possible sequences of partitions $\mathcal{P}$ and spans from $0$ (compositional) to $1$ (non-compositional).
\[
\text{re}(L, L^\star) = \min_{p \in \mathcal{P}} \text{re}(p, L, L^\star).
\]

\subsection{Results}
\label{sec:results}

Fig.~\ref{fig:master_plot} shows the main findings of our research. In plot (a), we see the parameter counts at the threshold. Below these values, the model cannot solve the task but above these, it can solve it. Further, observe the curve delineated by the lower left corner of the shift from unsuccessful to successful models. This inverse relationship between bits and parameters shows that the more parameters in the model, the fewer bits it needs to solve the task. Note however that it could only solve the task with fewer bits if it was forming a non-compositional code, suggesting that higher parameter models are able to do so while lower parameter ones cannot. 

Observe further that all of our models above the minimum threshold (72,400) have the capacity to learn a compositional code. This is shown by the perfect training accuracy achieved by all of those models in plot (a) for 24 bits and by the perfect compositionality (zero entropy) in plot (b) for 24 bits. Together with the above, this validates that learning compositional codes requires less capacity than learning non-compositional codes.

\begin{figure}[h]
    \vspace{-4mm}
    \centering
    \includegraphics[width=\linewidth,height=8cm]{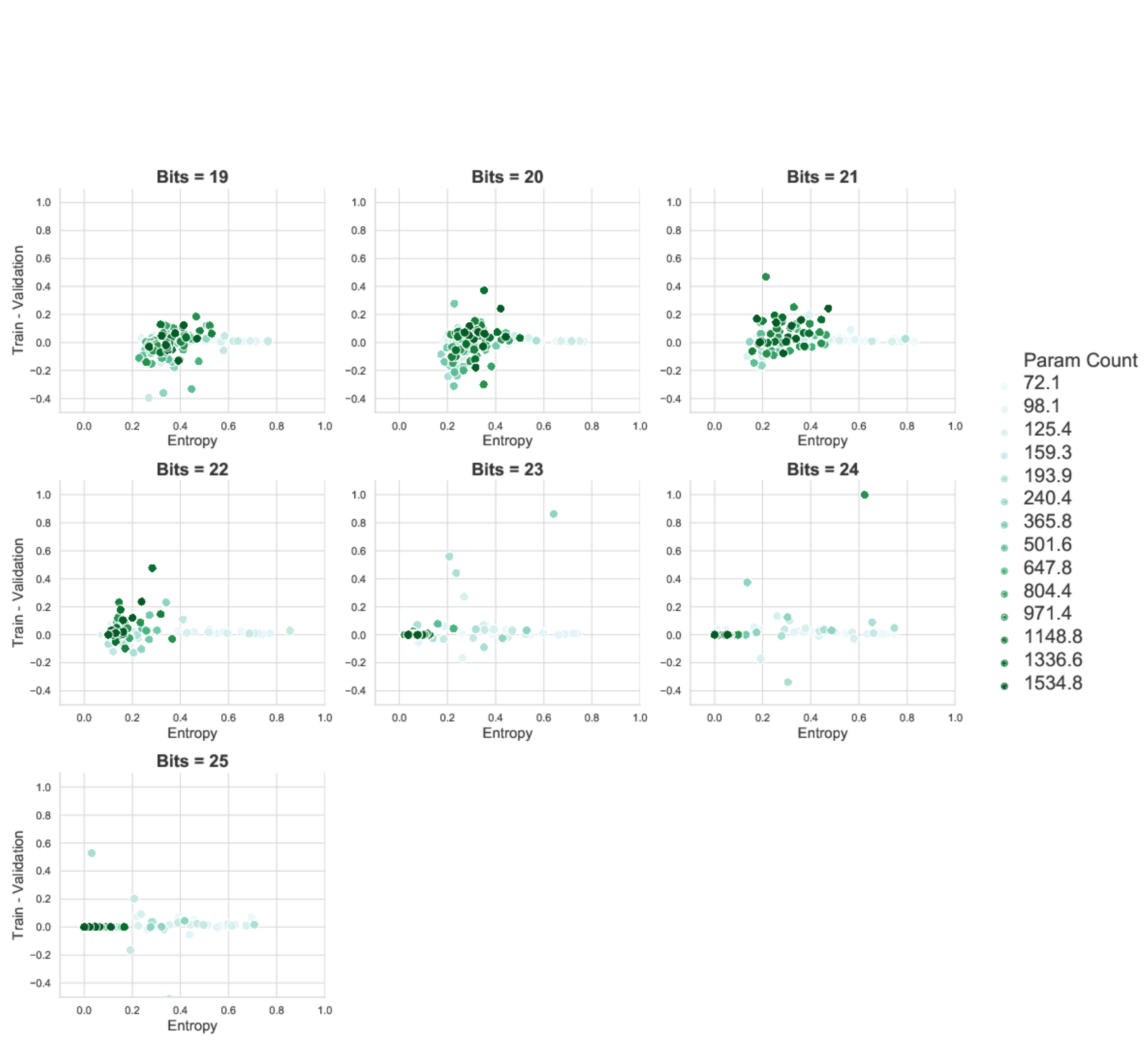}
    \caption{\textbf{Model A Entropy vs Overfitting}: Charts showing per-bit results for Entropy vs (Train - Validation) over the parameter range. 
    Observe the two models in bits $23$ and $24$ which were too successful in producing a non-compositional code and consequently overfit to the data.
    }
    \label{fig:trainvaldiff-entropy}
\end{figure}

Plot (c) confirms our hypothesis that large models can memorize the entire dataset. The 24 bit model with 971,400 parameters achieves a train accuracy of 1.0 and a validation accuracy of 0.0. Cross-validating this with plots (d) and (g), we find that a member of the same parameter class is non-compositional and that there is one that achieves unusually low recall. We verified that these are all the same seed, which shows that the agents in this model are memorizing the dataset. This is supplemented by Fig.~\ref{fig:trainvaldiff-entropy} where we see the non-compositional behavior by plotting residual entropy versus overfitting.

Plots (b) and (e) show that our compositionality metrics pass two sanity checks - high recall and perfect entropy can only be achieved with a channel that is sufficiently large (i.e. 24 bits) to allow for a compositional latent representation.

\begin{figure}[h]
    \centering
    \subfloat[Model A Training]{{\includegraphics[width=0.48\linewidth]{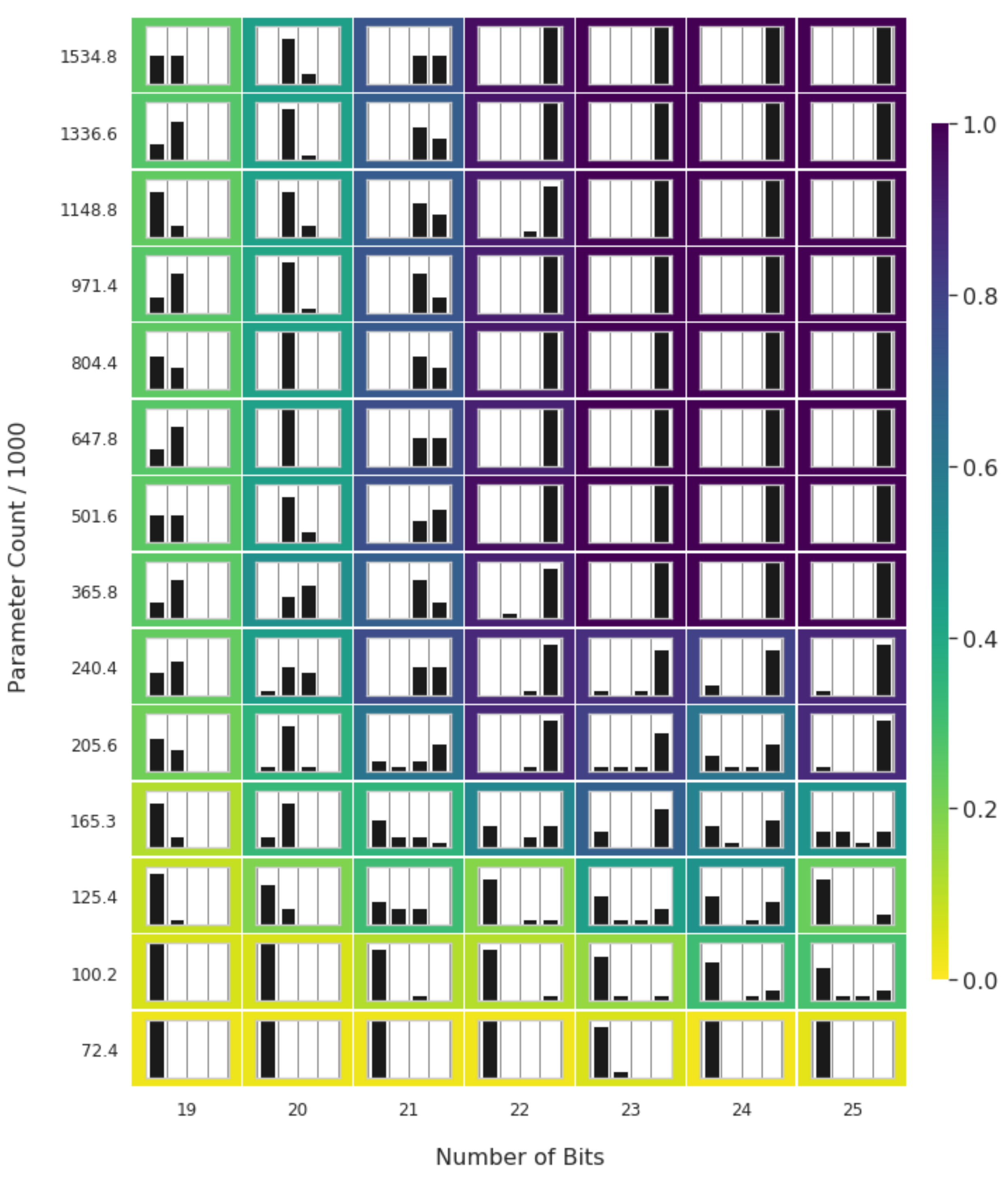}}}
    \hfill
    \subfloat[Model A Testing]{{\includegraphics[width=0.48\linewidth]{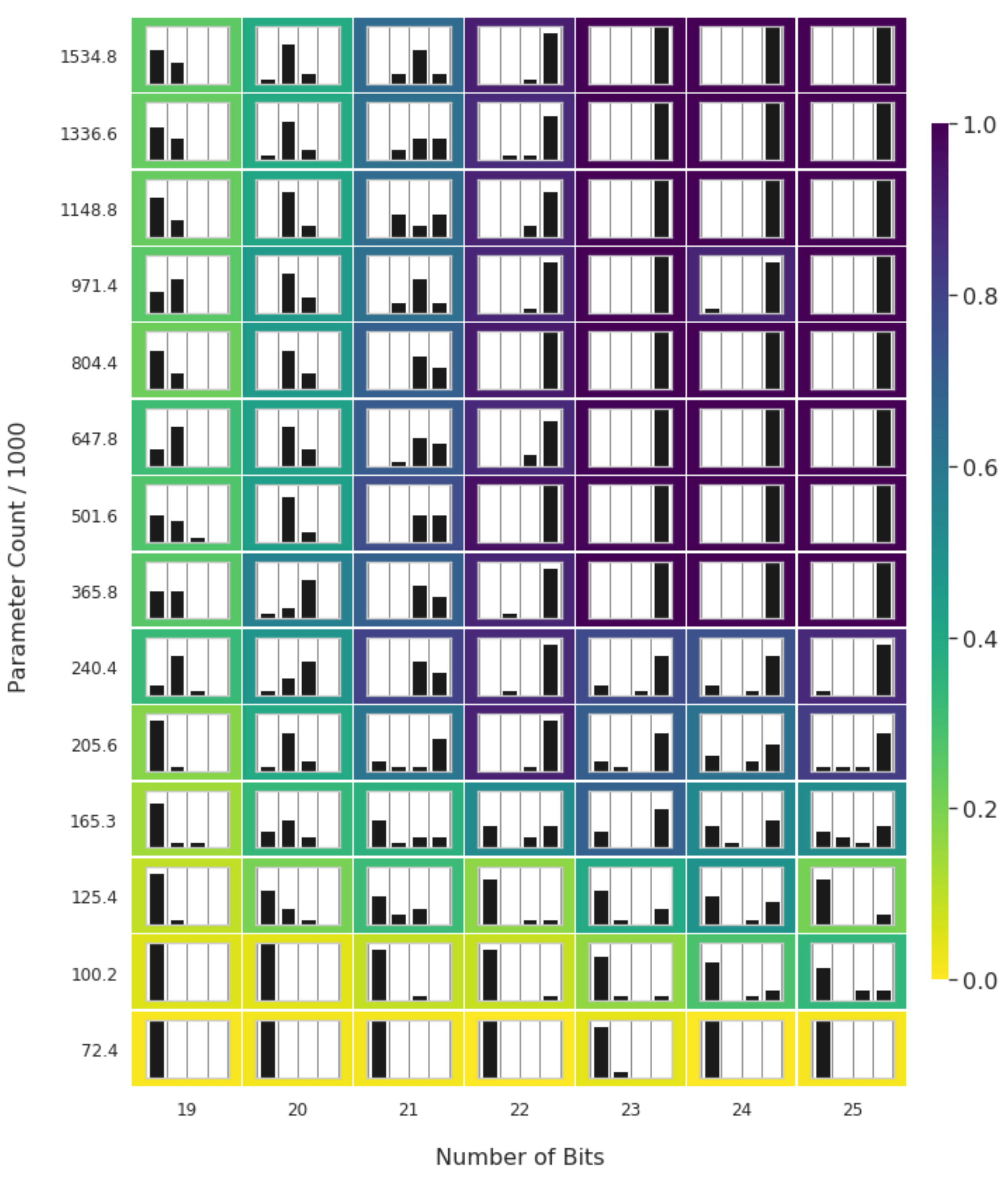}}}
    \hfill
    \subfloat[Model B Training]{{\includegraphics[width=0.48\linewidth]{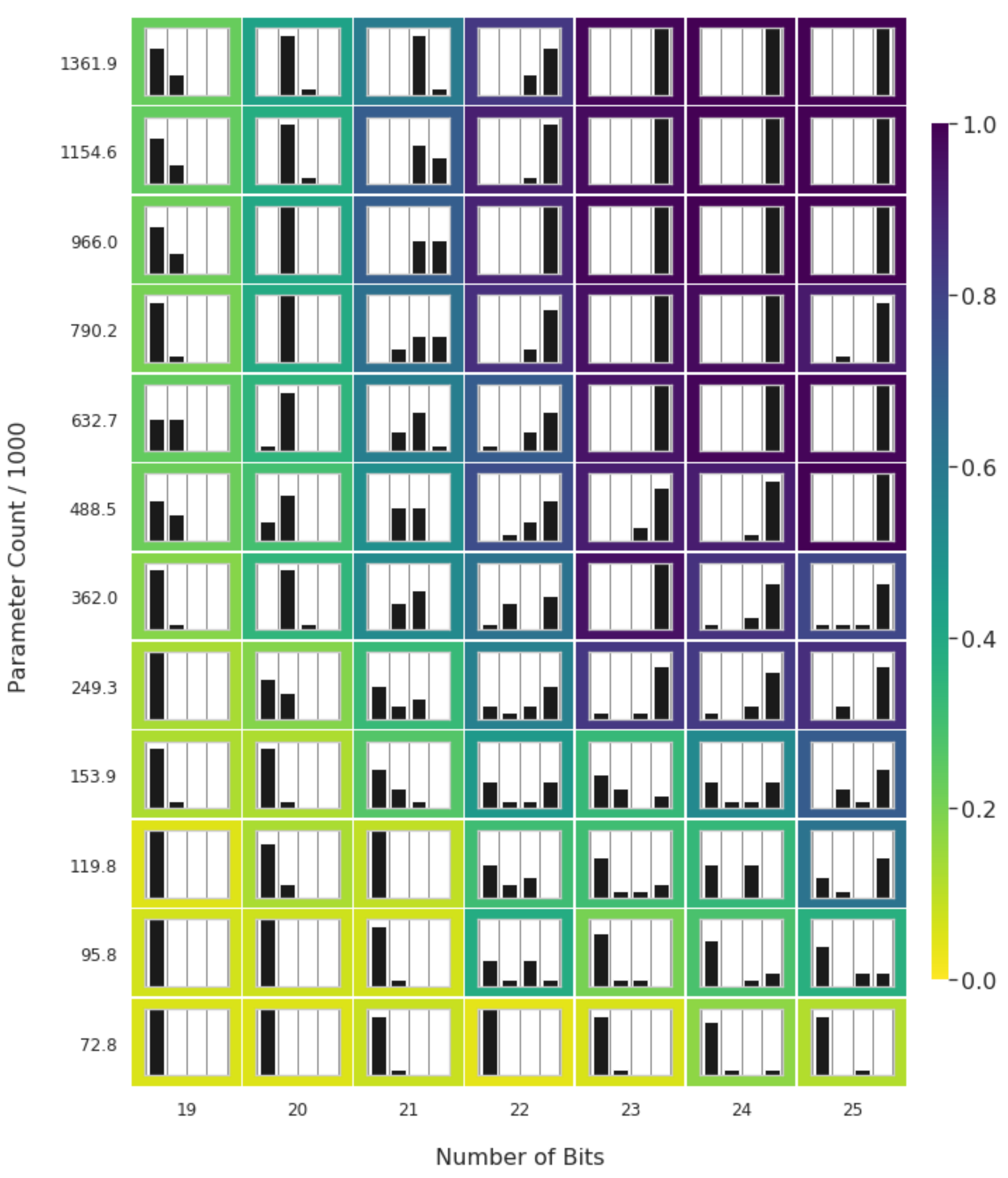}}}
    \hfill
    \subfloat[Model B Testing]{{\includegraphics[width=0.48\linewidth]{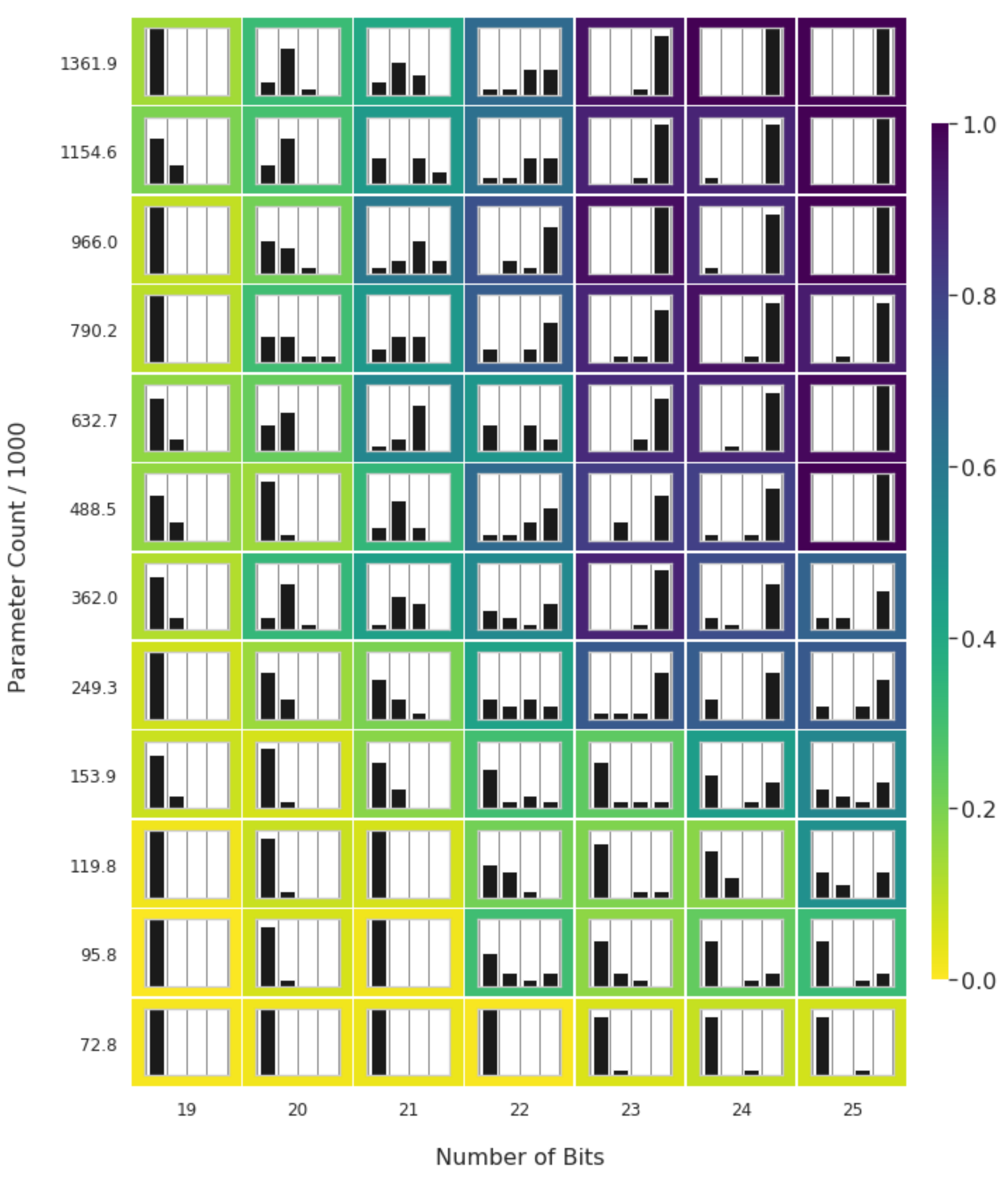}}}    
    \caption{Efficacy results for models A and B.} 
    \label{fig:efficacy}
\end{figure}

Plot (f) shows that while the capacity does not affect the ability to learn a compositional language across the model range, it does change the \emph{learnability}. Here we find that smaller models can fail to solve the task for any bandwidth, which coincides with literature suggesting a link between overparameterization and learnability \citep{li2018learning, du2018gradient}. This is supported by the efficacy results in Fig.~\ref{fig:efficacy}. 

Finally, as expected, we find that no model learns to solve the task with $< 20$ bits, validating that the minimum required number of bits for learning a language of size $|L|$ is $\lceil \log(|L|) \rceil$. We also see that no model learns to solve it for $20$ bits, which is likely due to optimization difficulties.

In Fig.~\ref{fig:recall-precision-entropy-A}, we present histograms showing precision, recall and residual entropy measured for each bit and parameter combination over the test set. The histograms show the distributions of these metrics, upon which we make a number of observations.

We first confirm the effectiveness of training by observing that almost all the models achieve perfect precision (Fig.~\ref{fig:recall-precision-entropy-A}~(a)), implying that $L \subseteq L^\star$, where $L$ is the language learned by the model. This occurs even with our learning objective in Eq.~\eqref{eq:variational-lowerbound} encouraging the model to capture all training strings rather than to focus on only a few training strings. 

A natural follow-up question is how large is $L^\star \backslash L$. We measure this with recall in Fig.~\ref{fig:recall-precision-entropy-A}~(b), which shows a clear phase transition according to the model capacity when $l \geq 22$. This agrees with what we saw in Fig.~\ref{fig:master_plot} and is equivalent to saying $| L^\star \backslash L| \gg 0$ at a value that is close to our predicted boundary of $l=\lceil \log_2 10^6 \rceil = 20$. We attribute this gap to the difficulty in learning a perfectly-parameterized neural network. 

In Fig.~\ref{fig:trainvaldiff-entropy} we show the empirical relation between entropy and overfitting over the parameter range. While it was hard for the models to overfit, there were some in bits 23 and 24 that did do so. For those models, the entropy was also found to be higher relative to the other models.

Even when $l \geq 22$, we observe training runs that fail to achieve optimal recall when the number of parameters is $\leq 365800$ (Fig.~\ref{fig:recall-precision-entropy-A}). Due to insufficient understanding of the relationship between the number of parameters and the capacity in a deep neural network, we cannot make a rigorous conclusion. We however conjecture that this is the upperbound to the minimal model capacity necessary to capture the tested compositional language. Above this threshold, the recall is almost always perfect, implying that the model has likely captured the compositional structure underlying $L^\star$ from a finite set of training strings. We run further experiments up to $1.5M$ parameters, but do not observe the expected overfitting. 

As shown in Fig.~\ref{fig:small-model}, we also run experiments with the number of categories reduced from $6$ to $4$ and similarly do not find the upperbound. It is left for future studies to determine why. Two conjectures that we have are that it is either due to insufficient model capacity to memorize the hash map between all the strings in $L^\star$ and $2^l$ latent strings or due to an inclination towards compositionality in our variational autoencoder.

These results clearly confirm the first part of our hypothesis - the latent sequence length must be at least as large as $\log |L^\star|$. They also confirm that there is a lowerbound on the number of parameters over which this model can successfully learn the underlying language. We have not been able to verify the upper bound in our experiments, which may require either a more (computationally) extensive set of experiments with even more parameters or a better theoretical understanding of the inherent biases behind learning with this architecture, such as from recent work on overparameterized models \citep{Belkin15849, nakkiran2020deep}.

\section{Conclusion}

In this paper, we hypothesize a thus far ignored connection between learnability, capacity, bandwidth, and compositionality for language learning. 
We empirically verfiy that learning the underlying compositional structure requires less capacity than memorizing a dataset. We also introduce a set of metrics to analyze the compositional properties of a learned language. These metrics are not only well motivated by theoretical insights, but are cross-validated by our task-specific metric. 

This paper opens the door for a vast amount of follow-up research. All our models were sufficiently large to represent the compositional structure of the language when given sufficient bandwidth, however there should be an upper bound for representing this compositional structure that we did not reach. We consider answering that to be the foremost question.

Furthermore, while large models did overfit, this was an exception rather than the rule. We hypothesize that this is due to the large number of examples in our language, which almost forces the model to generalize, but note that there are likely additional biases at play that warrant further investigation.

\appendix

\section*{Acknowledgements}

We would like to thank Marco Baroni and Angeliki Lazaridou for their comments on an earlier version of the paper. We would also like to thank the anonymous reviewers for giving insightful feedback in turn enhancing this work, particularly reviewer two for their thoroughness. Special thanks to Adam Roberts, Doug Eck, Mohammad Norouzi, and Jesse Engel.


\clearpage
\bibliographystyle{ACM-Reference-Format}  
\bibliography{refs}  


\begin{thebibliography}{00}


\ifx \showCODEN    \undefined \def \showCODEN     #1{\unskip}     \fi
\ifx \showDOI      \undefined \def \showDOI       #1{#1}\fi
\ifx \showISBNx    \undefined \def \showISBNx     #1{\unskip}     \fi
\ifx \showISBNxiii \undefined \def \showISBNxiii  #1{\unskip}     \fi
\ifx \showISSN     \undefined \def \showISSN      #1{\unskip}     \fi
\ifx \showLCCN     \undefined \def \showLCCN      #1{\unskip}     \fi
\ifx \shownote     \undefined \def \shownote      #1{#1}          \fi
\ifx \showarticletitle \undefined \def \showarticletitle #1{#1}   \fi
\ifx \showURL      \undefined \def \showURL       {\relax}        \fi
\providecommand\bibfield[2]{#2}
\providecommand\bibinfo[2]{#2}
\providecommand\natexlab[1]{#1}
\providecommand\showeprint[2][]{arXiv:#2}

\bibitem[\protect\citeauthoryear{Andreas}{Andreas}{2019}]%
        {andreas2018measuring}
\bibfield{author}{\bibinfo{person}{Jacob Andreas}.}
  \bibinfo{year}{2019}\natexlab{}.
\newblock \showarticletitle{Measuring Compositionality in Representation
  Learning}. In \bibinfo{booktitle}{{\em International Conference on Learning
  Representations}}.
\newblock
\showURL{%
\url{https://openreview.net/forum?id=HJz05o0qK7}}


\bibitem[\protect\citeauthoryear{Bahdanau, Murty, Noukhovitch, Nguyen,
  de~Vries, and Courville}{Bahdanau et~al\mbox{.}}{2019}]%
        {bahdanau2018systematic}
\bibfield{author}{\bibinfo{person}{Dzmitry Bahdanau}, \bibinfo{person}{Shikhar
  Murty}, \bibinfo{person}{Michael Noukhovitch}, \bibinfo{person}{Thien~Huu
  Nguyen}, \bibinfo{person}{Harm de Vries}, {and} \bibinfo{person}{Aaron
  Courville}.} \bibinfo{year}{2019}\natexlab{}.
\newblock \showarticletitle{Systematic Generalization: What Is Required and Can
  It Be Learned?}. In \bibinfo{booktitle}{{\em International Conference on
  Learning Representations}}.
\newblock
\showURL{%
\url{https://openreview.net/forum?id=HkezXnA9YX}}


\bibitem[\protect\citeauthoryear{Baroni}{Baroni}{2020}]%
        {baroni_linguistic_2019}
\bibfield{author}{\bibinfo{person}{Marco Baroni}.}
  \bibinfo{year}{2020}\natexlab{}.
\newblock \showarticletitle{Linguistic generalization and compositionality in
  modern artificial neural networks}.
\newblock \bibinfo{journal}{{\em Philosophical Transactions of the Royal
  Society B: Biological Sciences\/}}  \bibinfo{volume}{375} (\bibinfo{date}{02}
  \bibinfo{year}{2020}), \bibinfo{pages}{20190307}.
\newblock
\showDOI{%
\url{https://doi.org/10.1098/rstb.2019.0307}}


\bibitem[\protect\citeauthoryear{Barrett, Skyrms, and Cochran}{Barrett
  et~al\mbox{.}}{2018}]%
        {barrett_2018}
\bibfield{author}{\bibinfo{person}{Jeffrey~A. Barrett}, \bibinfo{person}{Brian
  Skyrms}, {and} \bibinfo{person}{Calvin Cochran}.}
  \bibinfo{year}{2018}\natexlab{}.
\newblock \showarticletitle{Hierarchical Models for the Evolution of
  Compositional Language}.
\newblock
\showURL{%
\url{http://philsci-archive.pitt.edu/14725/}}


\bibitem[\protect\citeauthoryear{Belkin, Hsu, Ma, and Mandal}{Belkin
  et~al\mbox{.}}{2019}]%
        {Belkin15849}
\bibfield{author}{\bibinfo{person}{Mikhail Belkin}, \bibinfo{person}{Daniel
  Hsu}, \bibinfo{person}{Siyuan Ma}, {and} \bibinfo{person}{Soumik Mandal}.}
  \bibinfo{year}{2019}\natexlab{}.
\newblock \showarticletitle{Reconciling modern machine-learning practice and
  the classical bias{\textendash}variance trade-off}.
\newblock \bibinfo{journal}{{\em Proceedings of the National Academy of
  Sciences\/}} \bibinfo{volume}{116}, \bibinfo{number}{32}
  (\bibinfo{year}{2019}), \bibinfo{pages}{15849--15854}.
\newblock
\showISSN{0027-8424}
\showDOI{%
\url{https://doi.org/10.1073/pnas.1903070116}}
\showeprint{https://www.pnas.org/content/116/32/15849.full.pdf}


\bibitem[\protect\citeauthoryear{Chaabouni, Kharitonov, Dupoux, and
  Baroni}{Chaabouni et~al\mbox{.}}{2019a}]%
        {chaabouni_anti-efficient_2019}
\bibfield{author}{\bibinfo{person}{Rahma Chaabouni}, \bibinfo{person}{Eugene
  Kharitonov}, \bibinfo{person}{Emmanuel Dupoux}, {and} \bibinfo{person}{Marco
  Baroni}.} \bibinfo{year}{2019}\natexlab{a}.
\newblock \showarticletitle{Anti-efficient encoding in emergent communication}.
\newblock In \bibinfo{booktitle}{{\em Advances in {Neural} {Information}
  {Processing} {Systems} 32}}, \bibfield{editor}{\bibinfo{person}{H.~Wallach},
  \bibinfo{person}{H.~Larochelle}, \bibinfo{person}{A.~Beygelzimer},
  \bibinfo{person}{F.~d{\textbackslash}textquotesingle Alché-Buc},
  \bibinfo{person}{E.~Fox}, {and} \bibinfo{person}{R.~Garnett}} (Eds.).
  \bibinfo{publisher}{Curran Associates, Inc.}, \bibinfo{pages}{6290--6300}.
\newblock
\showURL{%
\url{http://papers.nips.cc/paper/8859-anti-efficient-encoding-in-emergent-communication.pdf}}


\bibitem[\protect\citeauthoryear{Chaabouni, Kharitonov, Lazaric, Dupoux, and
  Baroni}{Chaabouni et~al\mbox{.}}{2019b}]%
        {chaabouni_word-order_2019}
\bibfield{author}{\bibinfo{person}{Rahma Chaabouni}, \bibinfo{person}{Eugene
  Kharitonov}, \bibinfo{person}{Alessandro Lazaric}, \bibinfo{person}{Emmanuel
  Dupoux}, {and} \bibinfo{person}{Marco Baroni}.}
  \bibinfo{year}{2019}\natexlab{b}.
\newblock \showarticletitle{Word-order biases in deep-agent emergent
  communication}.
\newblock \bibinfo{journal}{{\em arXiv:1905.12330 [cs]\/}} (\bibinfo{date}{May}
  \bibinfo{year}{2019}).
\newblock
\newblock
\shownote{arXiv: 1905.12330.}


\bibitem[\protect\citeauthoryear{Chen, Gilroy, Maletti, May, and Knight}{Chen
  et~al\mbox{.}}{2018}]%
        {chen_acl}
\bibfield{author}{\bibinfo{person}{Yining Chen}, \bibinfo{person}{Sorcha
  Gilroy}, \bibinfo{person}{Andreas Maletti}, \bibinfo{person}{Jonathan May},
  {and} \bibinfo{person}{Kevin Knight}.} \bibinfo{year}{2018}\natexlab{}.
\newblock \showarticletitle{Recurrent Neural Networks as Weighted Language
  Recognizers}. In \bibinfo{booktitle}{{\em Proceedings of the 2018 Conference
  of the North American Chapter of the Association for Computational
  Linguistics: Human Language Technologies, Volume 1 (Long Papers)}}.
  \bibinfo{publisher}{Association for Computational Linguistics},
  \bibinfo{pages}{2261--2271}.
\newblock
\showDOI{%
\url{https://doi.org/10.18653/v1/N18-1205}}


\bibitem[\protect\citeauthoryear{Collins, Sohl{-}Dickstein, and
  Sussillo}{Collins et~al\mbox{.}}{2017}]%
        {Collins_2017}
\bibfield{author}{\bibinfo{person}{Jasmine Collins}, \bibinfo{person}{Jascha
  Sohl{-}Dickstein}, {and} \bibinfo{person}{David Sussillo}.}
  \bibinfo{year}{2017}\natexlab{}.
\newblock \showarticletitle{Capacity and Trainability in Recurrent Neural
  Networks}. In \bibinfo{booktitle}{{\em International Conference on Learning
  Representations}}.
\newblock
\showURL{%
\url{https://openreview.net/forum?id=BydARw9ex}}


\bibitem[\protect\citeauthoryear{Du, Zhai, Poczos, and Singh}{Du
  et~al\mbox{.}}{2019}]%
        {du2018gradient}
\bibfield{author}{\bibinfo{person}{Simon~S. Du}, \bibinfo{person}{Xiyu Zhai},
  \bibinfo{person}{Barnabas Poczos}, {and} \bibinfo{person}{Aarti Singh}.}
  \bibinfo{year}{2019}\natexlab{}.
\newblock \showarticletitle{Gradient Descent Provably Optimizes
  Over-parameterized Neural Networks}. In \bibinfo{booktitle}{{\em
  International Conference on Learning Representations}}.
\newblock
\showURL{%
\url{https://openreview.net/forum?id=S1eK3i09YQ}}


\bibitem[\protect\citeauthoryear{Evtimova, Drozdov, Kiela, and Cho}{Evtimova
  et~al\mbox{.}}{2018}]%
        {evtimova2018emergent}
\bibfield{author}{\bibinfo{person}{Katrina Evtimova}, \bibinfo{person}{Andrew
  Drozdov}, \bibinfo{person}{Douwe Kiela}, {and} \bibinfo{person}{Kyunghyun
  Cho}.} \bibinfo{year}{2018}\natexlab{}.
\newblock \showarticletitle{Emergent Communication in a Multi-Modal, Multi-Step
  Referential Game}. In \bibinfo{booktitle}{{\em International Conference on
  Learning Representations}}.
\newblock
\showURL{%
\url{https://openreview.net/forum?id=rJGZq6g0-}}


\bibitem[\protect\citeauthoryear{Foerster, Assael, de~Freitas, and
  Whiteson}{Foerster et~al\mbox{.}}{2016}]%
        {foerster_learning_2016}
\bibfield{author}{\bibinfo{person}{Jakob Foerster},
  \bibinfo{person}{Ioannis~Alexandros Assael}, \bibinfo{person}{Nando de
  Freitas}, {and} \bibinfo{person}{Shimon Whiteson}.}
  \bibinfo{year}{2016}\natexlab{}.
\newblock \showarticletitle{Learning to Communicate with Deep Multi-Agent
  Reinforcement Learning}.
\newblock In \bibinfo{booktitle}{{\em Advances in Neural Information Processing
  Systems 29}}, \bibfield{editor}{\bibinfo{person}{D.~D. Lee},
  \bibinfo{person}{M.~Sugiyama}, \bibinfo{person}{U.~V. Luxburg},
  \bibinfo{person}{I.~Guyon}, {and} \bibinfo{person}{R.~Garnett}} (Eds.).
  \bibinfo{publisher}{Curran Associates, Inc.}, \bibinfo{pages}{2137--2145}.
\newblock
\showURL{%
\url{http://papers.nips.cc/paper/6042-learning-to-communicate-with-deep-multi-agent-reinforcement-learning.pdf}}


\bibitem[\protect\citeauthoryear{Havrylov and Titov}{Havrylov and
  Titov}{2017}]%
        {havrylov_emergence_2017}
\bibfield{author}{\bibinfo{person}{Serhii Havrylov} {and} \bibinfo{person}{Ivan
  Titov}.} \bibinfo{year}{2017}\natexlab{}.
\newblock \showarticletitle{Emergence of {Language} with {Multi}-agent {Games}:
  {Learning} to {Communicate} with {Sequences} of {Symbols}}.
\newblock In \bibinfo{booktitle}{{\em Advances in {Neural} {Information}
  {Processing} {Systems} 30}}, \bibfield{editor}{\bibinfo{person}{I.~Guyon},
  \bibinfo{person}{U.~V. Luxburg}, \bibinfo{person}{S.~Bengio},
  \bibinfo{person}{H.~Wallach}, \bibinfo{person}{R.~Fergus},
  \bibinfo{person}{S.~Vishwanathan}, {and} \bibinfo{person}{R.~Garnett}}
  (Eds.). \bibinfo{publisher}{Curran Associates, Inc.},
  \bibinfo{pages}{2149--2159}.
\newblock
\showURL{%
\url{http://papers.nips.cc/paper/6810-emergence-of-language-with-multi-agent-games-learning-to-communicate-with-sequences-of-symbols.pdf}}


\bibitem[\protect\citeauthoryear{Hochreiter and Schmidhuber}{Hochreiter and
  Schmidhuber}{1997}]%
        {Hochreiter:1997}
\bibfield{author}{\bibinfo{person}{Sepp Hochreiter} {and}
  \bibinfo{person}{J\"{u}rgen Schmidhuber}.} \bibinfo{year}{1997}\natexlab{}.
\newblock \showarticletitle{Long Short-Term Memory}.
\newblock \bibinfo{journal}{{\em Neural Computation\/}} \bibinfo{volume}{9},
  \bibinfo{number}{8} (\bibinfo{date}{Nov.} \bibinfo{year}{1997}),
  \bibinfo{pages}{1735–1780}.
\newblock
\showISSN{0899-7667}
\showDOI{%
\url{https://doi.org/10.1162/neco.1997.9.8.1735}}


\bibitem[\protect\citeauthoryear{Hupkes, Dankers, Mul, and Bruni}{Hupkes
  et~al\mbox{.}}{2020}]%
        {hupkes_compositionality_2019}
\bibfield{author}{\bibinfo{person}{Dieuwke Hupkes}, \bibinfo{person}{Verna
  Dankers}, \bibinfo{person}{Mathijs Mul}, {and} \bibinfo{person}{Elia Bruni}.}
  \bibinfo{year}{2020}\natexlab{}.
\newblock \showarticletitle{The compositionality of neural networks:
  integrating symbolism and connectionism}.
\newblock \bibinfo{journal}{{\em Journal of Artificial Intelligence
  Research\/}} (\bibinfo{year}{2020}).
\newblock


\bibitem[\protect\citeauthoryear{Jang, Gu, and Poole}{Jang
  et~al\mbox{.}}{2017}]%
        {jang_categorical_2016}
\bibfield{author}{\bibinfo{person}{Eric Jang}, \bibinfo{person}{Shixiang Gu},
  {and} \bibinfo{person}{Ben Poole}.} \bibinfo{year}{2017}\natexlab{}.
\newblock \showarticletitle{Categorical Reparameterization with
  Gumbel-Softmax}. In \bibinfo{booktitle}{{\em International Conference on
  Learning Representations}}.
\newblock
\showURL{%
\url{https://openreview.net/forum?id=rkE3y85ee}}


\bibitem[\protect\citeauthoryear{Kingma and Ba}{Kingma and Ba}{2015}]%
        {Kingma2015AdamAM}
\bibfield{author}{\bibinfo{person}{Diederik~P. Kingma} {and}
  \bibinfo{person}{Jimmy Ba}.} \bibinfo{year}{2015}\natexlab{}.
\newblock \showarticletitle{Adam: A Method for Stochastic Optimization}. In
  \bibinfo{booktitle}{{\em International Conference on Learning
  Representations}}.
\newblock


\bibitem[\protect\citeauthoryear{Kingma and Welling}{Kingma and
  Welling}{2014}]%
        {kingma2014auto}
\bibfield{author}{\bibinfo{person}{Diederik~P Kingma} {and}
  \bibinfo{person}{Max Welling}.} \bibinfo{year}{2014}\natexlab{}.
\newblock \showarticletitle{Auto-encoding Variational Bayes}. In
  \bibinfo{booktitle}{{\em International Conference on Learning
  Representations}}.
\newblock
\showURL{%
\url{https://openreview.net/forum?id=33X9fd2-9FyZd}}


\bibitem[\protect\citeauthoryear{Kirby, Tamariz, Cornish, and Smith}{Kirby
  et~al\mbox{.}}{2015}]%
        {kirbycompression}
\bibfield{author}{\bibinfo{person}{Simon Kirby}, \bibinfo{person}{Monica
  Tamariz}, \bibinfo{person}{Hannah Cornish}, {and} \bibinfo{person}{Kenny
  Smith}.} \bibinfo{year}{2015}\natexlab{}.
\newblock \showarticletitle{Compression and Communication in the Cultural
  Evolution of Linguistic Structure}.
\newblock \bibinfo{journal}{{\em Cognition\/}}  \bibinfo{volume}{141}
  (\bibinfo{year}{2015}), \bibinfo{pages}{87--102}.
\newblock
\showISSN{0010-0277}
\showDOI{%
\url{https://doi.org/10.1016/j.cognition.2015.03.016}}


\bibitem[\protect\citeauthoryear{Kottur, Moura, Lee, and Batra}{Kottur
  et~al\mbox{.}}{2017}]%
        {kottur2017natural}
\bibfield{author}{\bibinfo{person}{Satwik Kottur}, \bibinfo{person}{Jos{\'e}
  Moura}, \bibinfo{person}{Stefan Lee}, {and} \bibinfo{person}{Dhruv Batra}.}
  \bibinfo{year}{2017}\natexlab{}.
\newblock \showarticletitle{Natural Language Does Not Emerge `Naturally' in
  Multi-Agent Dialog}. In \bibinfo{booktitle}{{\em Proceedings of the 2017
  Conference on Empirical Methods in Natural Language Processing}}.
  \bibinfo{publisher}{Association for Computational Linguistics},
  \bibinfo{pages}{2962--2967}.
\newblock
\showDOI{%
\url{https://doi.org/10.18653/v1/D17-1321}}


\bibitem[\protect\citeauthoryear{Lake}{Lake}{2019}]%
        {lake_compgen}
\bibfield{author}{\bibinfo{person}{Brenden~M Lake}.}
  \bibinfo{year}{2019}\natexlab{}.
\newblock \showarticletitle{Compositional generalization through meta
  sequence-to-sequence learning}.
\newblock In \bibinfo{booktitle}{{\em Advances in Neural Information Processing
  Systems 32}}, \bibfield{editor}{\bibinfo{person}{H.~Wallach},
  \bibinfo{person}{H.~Larochelle}, \bibinfo{person}{A.~Beygelzimer},
  \bibinfo{person}{F.~d\textquotesingle Alch\'{e}-Buc},
  \bibinfo{person}{E.~Fox}, {and} \bibinfo{person}{R.~Garnett}} (Eds.).
  \bibinfo{publisher}{Curran Associates, Inc.}, \bibinfo{pages}{9788--9798}.
\newblock
\showURL{%
\url{http://papers.nips.cc/paper/9172-compositional-generalization-through-meta-sequence-to-sequence-learning.pdf}}


\bibitem[\protect\citeauthoryear{Lazaridou, Hermann, Tuyls, and
  Clark}{Lazaridou et~al\mbox{.}}{2018}]%
        {lazaridou_emergence_2018}
\bibfield{author}{\bibinfo{person}{Angeliki Lazaridou},
  \bibinfo{person}{Karl~Moritz Hermann}, \bibinfo{person}{Karl Tuyls}, {and}
  \bibinfo{person}{Stephen Clark}.} \bibinfo{year}{2018}\natexlab{}.
\newblock \showarticletitle{Emergence of Linguistic Communication from
  Referential Games with Symbolic and Pixel Input}. In \bibinfo{booktitle}{{\em
  International Conference on Learning Representations}}.
\newblock
\showURL{%
\url{https://openreview.net/forum?id=HJGv1Z-AW}}


\bibitem[\protect\citeauthoryear{Lazaridou, Peysakhovich, and Baroni}{Lazaridou
  et~al\mbox{.}}{2017}]%
        {lazaridou_multi-agent_2016}
\bibfield{author}{\bibinfo{person}{Angeliki Lazaridou},
  \bibinfo{person}{Alexander Peysakhovich}, {and} \bibinfo{person}{Marco
  Baroni}.} \bibinfo{year}{2017}\natexlab{}.
\newblock \showarticletitle{Multi-{Agent} {Cooperation} and the {Emergence} of
  ({Natural}) {Language}}. In \bibinfo{booktitle}{{\em International Conference
  on Learning Representations}}.
\newblock
\showURL{%
\url{https://openreview.net/forum?id=Hk8N3Sclg}}


\bibitem[\protect\citeauthoryear{Lee, Cho, and Kiela}{Lee
  et~al\mbox{.}}{2019}]%
        {lee_countering_2019}
\bibfield{author}{\bibinfo{person}{Jason Lee}, \bibinfo{person}{Kyunghyun Cho},
  {and} \bibinfo{person}{Douwe Kiela}.} \bibinfo{year}{2019}\natexlab{}.
\newblock \showarticletitle{Countering Language Drift via Visual Grounding}. In
  \bibinfo{booktitle}{{\em Proceedings of the 2019 Conference on Empirical
  Methods in Natural Language Processing and the 9th International Joint
  Conference on Natural Language Processing (EMNLP-IJCNLP)}}.
  \bibinfo{publisher}{Association for Computational Linguistics},
  \bibinfo{address}{Hong Kong, China}, \bibinfo{pages}{4376--4386}.
\newblock
\showDOI{%
\url{https://doi.org/10.18653/v1/D19-1447}}


\bibitem[\protect\citeauthoryear{Lee, Cho, Weston, and Kiela}{Lee
  et~al\mbox{.}}{2018}]%
        {lee_emergent_2017}
\bibfield{author}{\bibinfo{person}{Jason Lee}, \bibinfo{person}{Kyunghyun Cho},
  \bibinfo{person}{Jason Weston}, {and} \bibinfo{person}{Douwe Kiela}.}
  \bibinfo{year}{2018}\natexlab{}.
\newblock \showarticletitle{Emergent Translation in Multi-Agent Communication}.
  In \bibinfo{booktitle}{{\em International Conference on Learning
  Representations}}.
\newblock
\showURL{%
\url{https://openreview.net/forum?id=H1vEXaxA-}}


\bibitem[\protect\citeauthoryear{Lewis}{Lewis}{1969}]%
        {lewis1969convention}
\bibfield{author}{\bibinfo{person}{David Lewis}.}
  \bibinfo{year}{1969}\natexlab{}.
\newblock \bibinfo{booktitle}{{\em Convention: A philosophical study}}.
\newblock \bibinfo{publisher}{Harvard University Press}.
\newblock


\bibitem[\protect\citeauthoryear{Li and Liang}{Li and Liang}{2018}]%
        {li2018learning}
\bibfield{author}{\bibinfo{person}{Yuanzhi Li} {and} \bibinfo{person}{Yingyu
  Liang}.} \bibinfo{year}{2018}\natexlab{}.
\newblock \showarticletitle{Learning Overparameterized Neural Networks via
  Stochastic Gradient Descent on Structured Data}. In \bibinfo{booktitle}{{\em
  Advances in Neural Information Processing Systems 31}}.
  \bibinfo{publisher}{Curran Associates, Inc.}, \bibinfo{pages}{8157--8166}.
\newblock


\bibitem[\protect\citeauthoryear{Liska, Kruszewski, and Baroni}{Liska
  et~al\mbox{.}}{2018}]%
        {liska_2018}
\bibfield{author}{\bibinfo{person}{Adam Liska}, \bibinfo{person}{Germ{\'{a}}n
  Kruszewski}, {and} \bibinfo{person}{Marco Baroni}.}
  \bibinfo{year}{2018}\natexlab{}.
\newblock \showarticletitle{Memorize or generalize? Searching for a
  compositional {RNN} in a haystack}.
\newblock \bibinfo{journal}{{\em CoRR\/}}  \bibinfo{volume}{abs/1802.06467}
  (\bibinfo{year}{2018}).
\newblock
\showeprint[arxiv]{1802.06467}
\showURL{%
\url{http://arxiv.org/abs/1802.06467}}


\bibitem[\protect\citeauthoryear{Lowe*, Gupta*, Foerster, Kiela, and
  Pineau}{Lowe* et~al\mbox{.}}{2020}]%
        {lowe*2020on}
\bibfield{author}{\bibinfo{person}{Ryan Lowe*}, \bibinfo{person}{Abhinav
  Gupta*}, \bibinfo{person}{Jakob Foerster}, \bibinfo{person}{Douwe Kiela},
  {and} \bibinfo{person}{Joelle Pineau}.} \bibinfo{year}{2020}\natexlab{}.
\newblock \showarticletitle{On the interaction between supervision and
  self-play in emergent communication}. In \bibinfo{booktitle}{{\em
  International Conference on Learning Representations}}.
\newblock
\showURL{%
\url{https://openreview.net/forum?id=rJxGLlBtwH}}


\bibitem[\protect\citeauthoryear{Maddison, Mnih, and Teh}{Maddison
  et~al\mbox{.}}{2017}]%
        {maddison2016concrete}
\bibfield{author}{\bibinfo{person}{Chris~J. Maddison}, \bibinfo{person}{Andriy
  Mnih}, {and} \bibinfo{person}{Yee~Whye Teh}.}
  \bibinfo{year}{2017}\natexlab{}.
\newblock \showarticletitle{The Concrete Distribution: A Continuous Relaxation
  of Discrete Random Variables}. In \bibinfo{booktitle}{{\em International
  Conference on Learning Representations}}.
\newblock
\showURL{%
\url{https://openreview.net/forum?id=S1jE5L5gl}}


\bibitem[\protect\citeauthoryear{Mordatch and Abbeel}{Mordatch and
  Abbeel}{2018}]%
        {mordatch_emergence_2017}
\bibfield{author}{\bibinfo{person}{Igor Mordatch} {and} \bibinfo{person}{Pieter
  Abbeel}.} \bibinfo{year}{2018}\natexlab{}.
\newblock \showarticletitle{Emergence of Grounded Compositional Language in
  Multi-Agent Populations}. In \bibinfo{booktitle}{{\em AAAI Conference on
  Artificial Intelligence}}.
\newblock
\showURL{%
\url{https://aaai.org/ocs/index.php/AAAI/AAAI18/paper/view/17007}}


\bibitem[\protect\citeauthoryear{Nakkiran, Kaplun, Bansal, Yang, Barak, and
  Sutskever}{Nakkiran et~al\mbox{.}}{2020}]%
        {nakkiran2020deep}
\bibfield{author}{\bibinfo{person}{Preetum Nakkiran}, \bibinfo{person}{Gal
  Kaplun}, \bibinfo{person}{Yamini Bansal}, \bibinfo{person}{Tristan Yang},
  \bibinfo{person}{Boaz Barak}, {and} \bibinfo{person}{Ilya Sutskever}.}
  \bibinfo{year}{2020}\natexlab{}.
\newblock \showarticletitle{Deep Double Descent: Where Bigger Models and More
  Data Hurt}. In \bibinfo{booktitle}{{\em International Conference on Learning
  Representations}}.
\newblock
\showURL{%
\url{https://openreview.net/forum?id=B1g5sA4twr}}


\bibitem[\protect\citeauthoryear{Paszke, Gross, Massa, et~al\mbox{.}}{Paszke
  et~al\mbox{.}}{2019}]%
        {pytorch_NIPS2019}
\bibfield{author}{\bibinfo{person}{Adam Paszke}, \bibinfo{person}{Sam Gross},
  \bibinfo{person}{Francisco Massa}, {et~al\mbox{.}}}
  \bibinfo{year}{2019}\natexlab{}.
\newblock \showarticletitle{PyTorch: An Imperative Style, High-Performance Deep
  Learning Library}.
\newblock In \bibinfo{booktitle}{{\em NeurIPS}},
  \bibfield{editor}{\bibinfo{person}{H.~Wallach},
  \bibinfo{person}{H.~Larochelle}, \bibinfo{person}{A.~Beygelzimer},
  \bibinfo{person}{F.~d\textquotesingle Alch\'{e}-Buc},
  \bibinfo{person}{E.~Fox}, {and} \bibinfo{person}{R.~Garnett}} (Eds.).
  \bibinfo{publisher}{Curran Associates, Inc.}, \bibinfo{pages}{8024--8035}.
\newblock
\showURL{%
\url{http://papers.nips.cc/paper/9015-pytorch-an-imperative-style-high-performance-deep-learning-library.pdf}}


\bibitem[\protect\citeauthoryear{Spike, Stadler, Kirby, and Smith}{Spike
  et~al\mbox{.}}{2017}]%
        {Spike2017MinimalRF}
\bibfield{author}{\bibinfo{person}{Matthew Spike}, \bibinfo{person}{Kevin
  Stadler}, \bibinfo{person}{Simon Kirby}, {and} \bibinfo{person}{Kenny
  Smith}.} \bibinfo{year}{2017}\natexlab{}.
\newblock \showarticletitle{Minimal Requirements for the Emergence of Learned
  Signaling}. In \bibinfo{booktitle}{{\em Cognitive Science}}.
\newblock
\showURL{%
\url{https://doi.org/10.1111/cogs.12351}}


\bibitem[\protect\citeauthoryear{Steels}{Steels}{1997}]%
        {steels_1997}
\bibfield{author}{\bibinfo{person}{Luc Steels}.}
  \bibinfo{year}{1997}\natexlab{}.
\newblock \showarticletitle{The Synthetic Modeling of Language Origins}.
\newblock \bibinfo{journal}{{\em Evolution of Communication\/}}
  \bibinfo{volume}{1}, \bibinfo{number}{1} (\bibinfo{year}{1997}),
  \bibinfo{pages}{1--34}.
\newblock
\showISSN{1387-5337}
\showDOI{%
\url{https://doi.org/10.1075/eoc.1.1.02ste}}


\bibitem[\protect\citeauthoryear{Steels and Kaplan}{Steels and Kaplan}{2000}]%
        {Aibo}
\bibfield{author}{\bibinfo{person}{Luc Steels} {and}
  \bibinfo{person}{Frédéric Kaplan}.} \bibinfo{year}{2000}\natexlab{}.
\newblock \showarticletitle{AIBO’s first words: The social learning of
  language and meaning}.
\newblock \bibinfo{journal}{{\em Evolution of Communication\/}}
  \bibinfo{volume}{4} (\bibinfo{year}{2000}).
\newblock
\showURL{%
\url{http://www.jbe-platform.com/content/journals/10.1075/eoc.4.1.03ste}}


\bibitem[\protect\citeauthoryear{Sukhbaatar, Szlam, and Fergus}{Sukhbaatar
  et~al\mbox{.}}{2016}]%
        {sukhbaatar2016learning}
\bibfield{author}{\bibinfo{person}{Sainbayar Sukhbaatar},
  \bibinfo{person}{Arthur Szlam}, {and} \bibinfo{person}{Rob Fergus}.}
  \bibinfo{year}{2016}\natexlab{}.
\newblock \showarticletitle{Learning Multiagent Communication with
  Backpropagation}.
\newblock In \bibinfo{booktitle}{{\em NeurIPS}},
  \bibfield{editor}{\bibinfo{person}{D.~D. Lee}, \bibinfo{person}{M.~Sugiyama},
  \bibinfo{person}{U.~V. Luxburg}, \bibinfo{person}{I.~Guyon}, {and}
  \bibinfo{person}{R.~Garnett}} (Eds.). \bibinfo{publisher}{Curran Associates,
  Inc.}, \bibinfo{pages}{2244--2252}.
\newblock
\showURL{%
\url{http://papers.nips.cc/paper/6398-learning-multiagent-communication-with-backpropagation.pdf}}


\bibitem[\protect\citeauthoryear{Verhoef, Kirby, and de~Boer}{Verhoef
  et~al\mbox{.}}{2016}]%
        {tessa_2016}
\bibfield{author}{\bibinfo{person}{Tessa Verhoef}, \bibinfo{person}{Simon
  Kirby}, {and} \bibinfo{person}{Bart de Boer}.}
  \bibinfo{year}{2016}\natexlab{}.
\newblock \showarticletitle{Iconicity and the Emergence of Combinatorial
  Structure in Language}.
\newblock \bibinfo{journal}{{\em Cognitive Science\/}} \bibinfo{volume}{40},
  \bibinfo{number}{8} (\bibinfo{year}{2016}), \bibinfo{pages}{1969--1994}.
\newblock
\showISSN{1551-6709}
\showDOI{%
\url{https://doi.org/10.1111/cogs.12326}}


\bibitem[\protect\citeauthoryear{Weber, Shekhar, and Balasubramanian}{Weber
  et~al\mbox{.}}{2018}]%
        {weber_2018}
\bibfield{author}{\bibinfo{person}{Noah Weber}, \bibinfo{person}{Leena
  Shekhar}, {and} \bibinfo{person}{Niranjan Balasubramanian}.}
  \bibinfo{year}{2018}\natexlab{}.
\newblock \showarticletitle{The Fine Line between Linguistic Generalization and
  Failure in Seq2Seq-Attention Models}. In \bibinfo{booktitle}{{\em Proceedings
  of the Workshop on Generalization in the Age of Deep Learning}}.
  \bibinfo{publisher}{Association for Computational Linguistics},
  \bibinfo{pages}{24--27}.
\newblock
\showDOI{%
\url{https://doi.org/10.18653/v1/W18-1004}}


\bibitem[\protect\citeauthoryear{Zaslavsky, Kemp, Regier, and Tishby}{Zaslavsky
  et~al\mbox{.}}{2018}]%
        {zaslavsky_efficient_2018}
\bibfield{author}{\bibinfo{person}{Noga Zaslavsky}, \bibinfo{person}{Charles
  Kemp}, \bibinfo{person}{Terry Regier}, {and} \bibinfo{person}{Naftali
  Tishby}.} \bibinfo{year}{2018}\natexlab{}.
\newblock \showarticletitle{Efficient compression in color naming and its
  evolution}.
\newblock \bibinfo{journal}{{\em Proceedings of the National Academy of
  Sciences\/}} \bibinfo{volume}{115}, \bibinfo{number}{31}
  (\bibinfo{date}{July} \bibinfo{year}{2018}), \bibinfo{pages}{7937--7942}.
\newblock
\showISSN{0027-8424, 1091-6490}
\showDOI{%
\url{https://doi.org/10.1073/pnas.1800521115}}


\end{thebibliography}

\end{document}